\documentclass[journal]{IEEEtran}
\usepackage{times}

\usepackage{amsmath,amsfonts,amsthm,amssymb}
\usepackage{algorithmic}
\usepackage{algorithm}
\usepackage{array}
\usepackage{textcomp}
\usepackage{stfloats}
\usepackage{url}
\usepackage{verbatim}
\usepackage{graphicx}
\usepackage{subcaption}
\usepackage{cite}
\usepackage{color}
\usepackage{svg}
\usepackage{xspace}
\usepackage{authblk}
\usepackage{booktabs}
\usepackage{eso-pic}

\usepackage{adjustbox}
\usepackage{tikz}
\usetikzlibrary{automata,positioning,shapes,arrows,shadows,shadows.blur}
\newcommand{\shadowString}{circular drop shadow}

\usepackage[intoc]{nomencl}

\DeclareMathOperator*{\argmin}{\arg\!\min}

\usepackage[numbers]{natbib}
\usepackage{multicol}
\usepackage[bookmarks=true]{hyperref}

\begin{document}

\AddToShipoutPictureBG*{%
  \AtPageUpperLeft{%
    \hspace{16.5cm}%
    \raisebox{-1.5cm}{%
      \makebox[0pt][r]{Preprint submitted to the IEEE Transactions on Robotics journal.}}}}
      
\title{
Online Pareto-Optimal Decision-Making for Complex Tasks using Active Inference
}


\author{
Peter Amorese\textsuperscript{\dag}, 
Shohei Wakayama\textsuperscript{\dag}\thanks{\textsuperscript{\dag}Authors contributed equally to this work}, 
Nisar Ahmed, 
and Morteza Lahijanian
\thanks{Authors are with the University of Colorado Boulder.}
}

\maketitle

\newtheorem{definition}{Definition}
\newtheorem{problem}{Problem}
\newtheorem{example}{Example}
\newtheorem{remark}{Remark}
\newtheorem{lemma}{Lemma}

\newcommand{\pa}[1]{\textcolor{magenta}{[PA: #1]}}
\newcommand{\sw}[1]{\textcolor{red}{[SW: #1]}}
\newcommand{\ml}[1]{\textcolor{blue}{[ML: #1]}}
\newcommand{\na}[1]{\textcolor{red}{[NA: #1]}}
\newcommand{\qh}[1]{\textcolor{magenta}{[QH: #1]}}
\newcommand{\zl}[1]{\textcolor{cyan}{[ZL: #1]}}

\newcommand{\edt}[1]{\textcolor{black}{#1}}

\newcommand{\reals}{\mathbb{R}}
\newcommand{\naturals}{\mathbb{N}}
\newcommand{\N}{\mathcal{N}}
\newcommand{\C}{\mathcal{C}}
\renewcommand{\L}{\mathcal{L}}
\newcommand{\needcite}{\textbf{CITE} } 
\newcommand{\dist}{\mathcal{D}}
\newcommand{\scurr}{s_K}
\newcommand{\distest}{\tilde{\mathcal{D}}}
\newcommand{\expec}[1]{\mathbb{E}[#1]}
\newcommand{\variance}[1]{\text{var}[#1]}
\newcommand{\entropy}[1]{\mathbf{H}[#1]}
\newcommand{\st}{\text{ s.t. }}
\newcommand{\cvec}{\mathbf{c}}
\newcommand{\ccvec}{\mathbf{C}}
\newcommand{\cvecest}{\tilde{\mathbf{c}}}
\newcommand{\ccvecest}{\tilde{\mathbf{C}}}
\newcommand{\kld}{D_{KL}}
\newcommand{\planparams}{\theta_{s_c, \pi^*}}
\newcommand{\planset}{A^*}
\newcommand{\actiontxt}[1]{\texttt{#1}}

\newcommand{\mean}{\mathbf{\mu}}
\newcommand{\cov}{\mathbf{\Sigma}}

\newcommand{\DTS}{\text{DTS-sc}\xspace}
\newcommand{\pref}{\mathrm{pr}}

\newcommand{\Q}{{Q_\phi}}

\newcommand{\Rone}{\textcolor{teal}{\textbf{R1}}}
\newcommand{\Rtwo}{\textcolor{violet}{\textbf{R2}}}
\newcommand{\Rthree}{\textcolor{blue}{\textbf{R3}}}

\begin{abstract}
When a robot autonomously performs a complex task, it frequently must balance competing objectives while maintaining safety. This becomes more difficult in uncertain environments
with stochastic outcomes.
Enhancing transparency in the robot's behavior and aligning with user preferences are also crucial. 
This paper introduces a novel framework for multi-objective reinforcement learning that ensures safe task execution, optimizes trade-offs between objectives, and adheres to user preferences.
The framework has two main layers: a multi-objective task planner and a high-level selector. 
The planning layer generates a set of optimal trade-off plans that guarantee satisfaction of a temporal logic task.
The selector uses active inference to decide which generated plan best complies with user preferences and aids learning. 
Operating iteratively, the framework updates a parameterized learning model based on collected data.
Case studies and benchmarks on both manipulation and mobile robots show that our framework outperforms other methods and (i) learns multiple optimal trade-offs, (ii) adheres to a user preference, and (iii) allows the user to adjust the balance between (i) and (ii).
\end{abstract}


\IEEEpeerreviewmaketitle


\section{Introduction} \label{sec: introduction}

\IEEEPARstart{T}{he} demand for robots to autonomously perform hazardous and repetitive tasks is steadily increasing~\cite{gao2017,kim2019control}. Such tasks often involve multiple, possibly competing, quantitative objectives, e.g., time and energy, requiring optimization for trade-offs between the objectives, known as {\it Pareto optimal}.  
In unknown environments, like the Martian surface or civilian households, however, such quantities are often stochastic and unknown {\it a priori}. 
Fortunately, they are measurable and can be learned online. 
Then, the robot should make decisions to learn \textit{all} optimal trade-offs, or focus on learning a single \textit{user-preferred} trade-off. Although highly desirable, accomplishing both simultaneously is challenging since the trade-off selection must balance exploring unknown trade-offs with exploiting preferred trade-offs. This paper focuses on this challenge and aims to develop a framework that simultaneously (i) learns multiple optimal trade-offs, (ii) embeds an intuitive user preference over a desired trade-off, and (iii) allows the user to adjust the balance between (i) and (ii).

\begin{figure}[t]
    \centering
    \includegraphics[width=0.55\columnwidth]{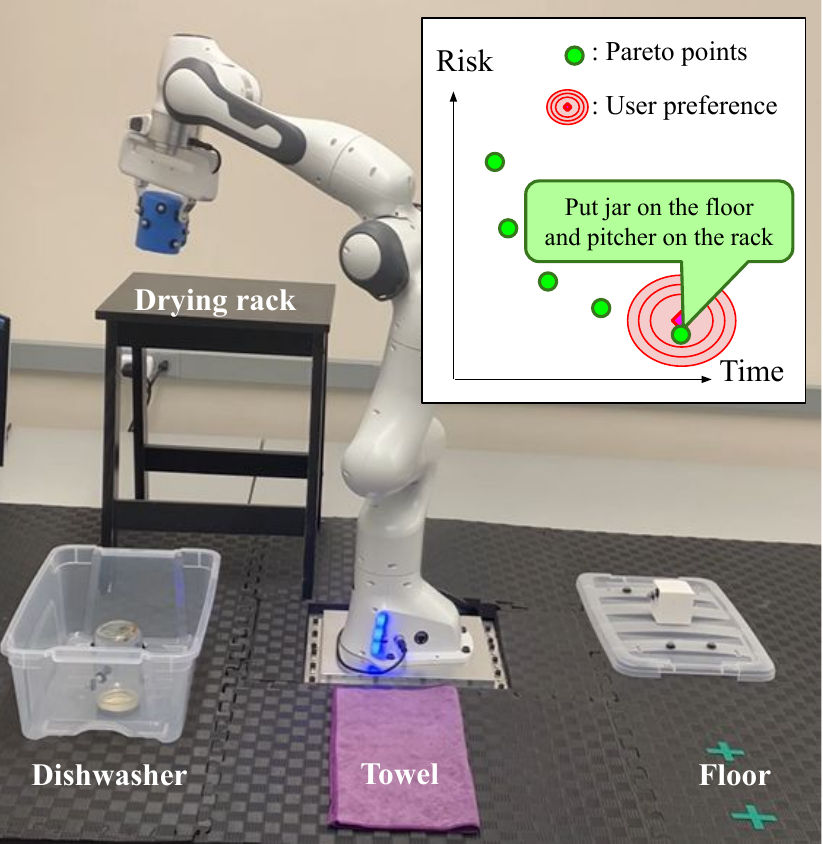}
    \caption{Motivating robotic dishwashing scenario: a robotic manipulator needs to compare execution time of a task and inherent risks of dropping fragile dishes (the jar is more fragile than the blue pitcher). The user's preferred trade-off between time and risk is incorporated for transparent behaviors.}
    \label{fig: scenario}
    \vspace{-0.2in}
\end{figure}

\begin{example} \label{ex: scenario}
\emph{
Consider the
robotic dishwashing scenario in Figure~\ref{fig: scenario}. 
The robotic manipulator is equipped with 
a sampling-based motion planner (e.g., RRT and PRM \cite{kavraki1996probabilistic,karaman2011sampling}) that solves A-to-B planning queries. 
The robot must repetitively complete a complex task of inserting the dishes into the dishwasher, then properly drying them by either placing directly up on the rack, or placing on the towel and moving them to the floor.
In such a scenario, the robot should be both efficient (minimize the time of execution) and risk-aware when carrying fragile dishes high above the ground.
However, due to the nature of sampling-based methods, the characteristics of actions (time and path-height) are unknown {\it a priori}, and hence the robot needs to construct accurate models of these quantities during deployment. 
The robot can either quickly place dishes up on a drying rack (low-time, high-risk for fragile objects), or manually dry dishes on the ground (high-time, low-risk). 
Considering a human user may prefer one trade-off over the other,
the robot must both learn the quantitative models, and decide amongst optimal trade-off actions while adhering to the user's preference. Such decision making problems involving competing objectives can be formulated as multi-objective reinforcement learning (MORL) \cite{wang2012,abels2019,hayes2022,mirzanejad2022}.
}
\end{example}


Embedding the satisfaction of a task into the multi-objective reward function often aims to capture both qualitative properties of the task (i.e. whether or not the robot completed the task), as well as quantitative properties (i.e. how well the robot completed the task) \cite{haarnoja2018soft}. However, properly representing qualitative task completion through a reward function is a challenging problem, often requiring the use of methods such as inverse reinforcement learning (IRL) \cite{arora2021survey}. In uncertain environments, it is not always possible to obtain large amounts of human teaching data. Furthermore, these learning-based methods often fail to provide safety guarantees on the qualitative behavior of the robot.
Nevertheless, in scenarios where qualitative aspects of the robotic model and task are known beforehand, e.g., manipulation problems, {\it formal methods} 
\cite{Lahijanian:AR-CRAS:2018}
provide alternative approaches to both task representation and plan
synthesis using temporal logic specifications, such as Linear Temporal Logic (LTL) \cite{baier2008} and LTL over finite traces (LTLf) \cite{de2013linear}. 
These planning methods can solve the qualitative portion of the learning problem and provide guarantees on the completion of tasks and safety while avoiding the need for IRL. 
Formal synthesis approaches have also been used to solve multi-objective quantitative planning problems \cite{amorese2023}; however, those approaches assume a perfectly accurate quantitative model, 
which are often unavailable.


Unlike single-objective RL, MORL agents must select among potentially many optimal trade-offs. The numerical valuations of these optimal trade-offs are referred to as the Pareto front.
The Multi-Objective Multi-Armed Bandit \edt{(MOMAB)} community emphasizes learning all optimal trade-off actions, yielding valuable insight into the trade-off analysis of the learned results \cite{drugan2013, busa-fekete2017, turgay2018, lu2019}. Alternatively, one can use \textit{scalarization} to convert the multi-objective problem into a single objective problem based on a \textit{preference} trade-offs \cite{wang2012, abels2019, hayes2022, mirzanejad2022}. Further description of these methods can be found in Sec. \ref{sec: background}.
To the best of our knowledge, no MORL selection methods simultaneously propose mathematically sound adherence to a \textit{preferred} trade-off while learning and exploring portions of the true Pareto front.

In light of these gaps, this study proposes a novel MORL framework 
centered around viewing optimal trade-off selection as its own high-level decision making under uncertainty problem.
Specifically, we employ \textit{active inference} (AIF),
recently explored as a sound approach for sequential decision-making under uncertainty \cite{friston2015,kaplan2018,friston2022, wakayama2023} that minimizes an information-theoretic quantity known as \textit{surprise}.
We exploit the decision making power of AIF to adhere to a user's ideal trade-off while exploring and learning localized portions of the whole Pareto front. This, however, introduces intractable computational complexity in optimizing for surprise for finite-horizon tasks.
Our framework integrates multi-objective \textit{planning} for a complex LTLf specification with high-level \textit{selection} of an optimal trade-off by using a Bayesian approach to learning the quantitative model. We extend AIF to reasoning over finite horizon planning via upper-bounding the surprise with \textit{expected free energy} (EFE). We derive computationally tractable techniques for optimizing EFE for LTLf tasks.
Using our framework, we  
illustrate the performance trade-offs between biasing towards a user's preferred Pareto point versus learning the entire Pareto front. Additionally, we showcase the utility of our approach using robotic case studies, including a hardware demonstration of the scenario in Example~\ref{ex: scenario}.

The major contributions of this work are three-fold:
\begin{itemize}
    \item A novel MORL framework that combines formally guaranteed task plan synthesis with efficient learning of Pareto-optimal behavior,
    \item Derivation of a tractable approximation of a \textit{free energy} formulation for finite horizon planning over an uncertain quantitative model,
    \item Benchmarking the efficacy of the proposed AIF-based selection strategy against the state-of-the-art methods \cite{abels2019, lu2019, mirzanejad2022}, along with Mars surface exploration simulation and a robotic hardware diswashing experiment.
\end{itemize}

A detailed discussion on related work along with an overview of AIF are provided in Sec.~\ref{sec: background}. 

\section{Preliminaries and Problem Formulation} \label{sec: problem_formulation}
This paper studies a framework that enables a robot to (i) repetitively complete a given high-level complex task, and (ii) learn the most efficient and user-aligned means of completing the task. These two goals separate nicely into \textit{qualitative} behavior, i.e., whether or not the robot completes the task, and \textit{quantitative} behavior, i.e., how well the robot completes the task. We restrict our attention to problems where qualitative aspects of the model (i.e. robotic states, state transitions, task related observations, etc.) are known, but quantitative aspects (i.e. costs of executing robotic actions) are unknown and must be learned. Below, we first formally define a robot model 
and LTLf task specifications.
We then connect these definitions with cost uncertainty and user preferences to be able to fully formulate the problem. 


%
%
%


\subsection{Robot Model} \label{sec: robotic_model}
\edt{We consider a robotic system modeled as a deterministic transition system (DTS), an abstraction used in many formal approaches to robotic systems, including both mobile robots and robotic manipulators \cite{kloetzer2008fully, fainekos2005temporal, yang2020secure, fainekos2005hybrid, he2015towards, He:RAL:2019, Lahijanian:AR-CRAS:2018}.}
To capture uncertainty in quantitative values, we augment classical DTS with a stochastic cost function as defined below.
\begin{definition}[\DTS]
    A labeled \emph{Deterministic Transition System} with a stochastic cost function (\DTS) is a tuple $T = (S, A, \delta_T, \mathcal{C}, AP, L)$, where
    \begin{itemize}
        \item $S$ is a finite set of states,
        \item $A$ is a finite set of actions,
        \item $\delta_T : S \times A \mapsto S$ is a deterministic transition function,
        \item $\mathcal{C} : S \times A \mapsto \mathcal{D}(\mathbb{R}^N)$ is a stochastic cost function that maps each state-action pair to an $N$-dimensional probability distribution, where $N \in \mathbb{N}$ and $\mathcal{D}(\mathbb{R}^N)$ is the set of all probability distributions over $\mathbb{R}^N$,
        \item $AP$ is a set of atomic propositions related to the  robot tasks, and
        \item $L : S \mapsto 2^{AP}$ is a labeling function that assigns to each state $s \in S$ the subset of propositions in $AP$ that are true in $s$.
    \end{itemize}
    \label{def:ts}
\end{definition}
\noindent

A robot plan $\pi \in A^*$ is a finite sequence of actions $\pi\!=\!a_0 a_1 \ldots a_m$ that induces a trajectory $\tau(s_0, \pi)\!=\!s_0 s_1 \ldots s_m s_{m+1}$ 
from a current state $s_0\!\in\!S$,  where $s_k\!\in\!S$ and, for all $0 \leq k \leq m+1$, $s_{k+1} = \delta_T(s_k, a_k)$.
The observation of $\tau(s_0,\pi)$, denoted $L(s_0,\pi)$, is a trace $L(s_0,\pi) = o_0 o_1 \ldots o_{m+1}$ where $o_k = L(s_k)$.

\begin{figure}[t]
    \centering
    \begin{adjustbox}{scale=0.7}
    \centering
    \begin{tikzpicture}[shorten >=1pt,node distance=6cm,>=stealth',thick, auto,
                        every state/.style={fill,very thick,black!20,text=black,\shadowString},
                        robot/.style = {fill,very thick,black!20,rounded corners, text=black, shape=rectangle, minimum height=1cm,minimum width=1cm, blur shadow},
                        accepting/.style ={blue!50!black!50,text=white,accepting by double},
                        initial/.style ={red!80!black!40,text=black,initial by arrow, initial left}, initial text=$ $]
                        
        \node[robot, label=below:\texttt{$\{wash\}$}] (s5) {$J_d, P_d$};
        \node[robot, initial, label=above:\texttt{$\{dry\}$}] (s1) [below left=0.3cm and 2.5cm of s5] {$J_f, P_f$};
        \node[robot, label=above:\texttt{$\{dry\}$}] (s2) [above left=2.1cm and 1.5cm of s5] {$J_r, P_f$};
        \node[robot, label=above:\texttt{$\{dry\}$}] (s3) [above right=2.1cm and 1.5cm of s5] {$J_r, P_r$};
        \node[robot, label=above:\texttt{$\{dry\}$}] (s4) [below right=0.3cm and 2.5cm of s5] {$J_f, P_r$};

        \path[->]
           (s1) edge [bend right, below, red] node [black, below=-0.2cm] {
           \includegraphics[width=1.0cm]{ 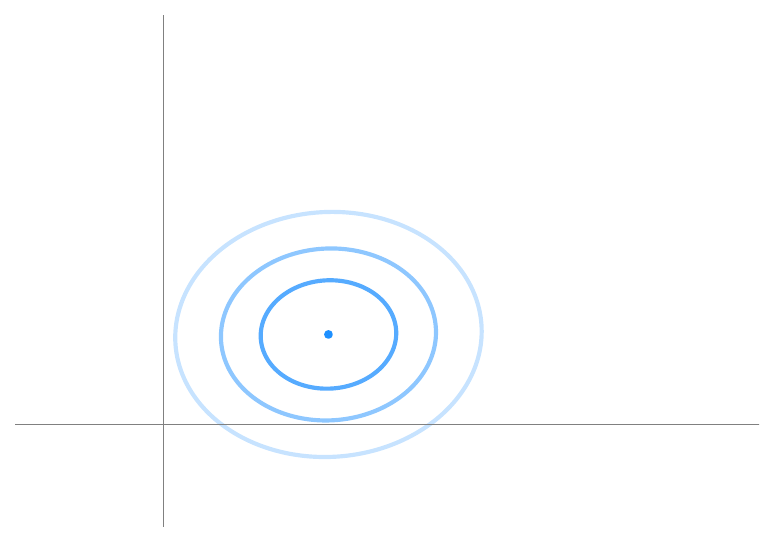} 
           } node [black, above] {\actiontxt{load}} (s5)
           
           (s5) edge [bend right, below, black] node [black, below=-0.2cm] {
           \includegraphics[width=1.0cm]{ 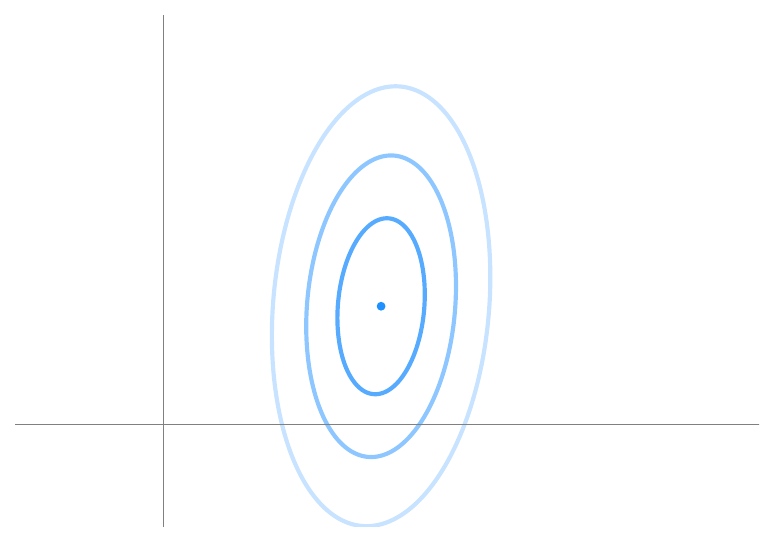} 
           } node [black, above] {\actiontxt{unload\_1}} (s1)

           (s2) edge [bend right, below, black] node [black, below=-0.2cm] {
           \includegraphics[width=1.0cm]{ 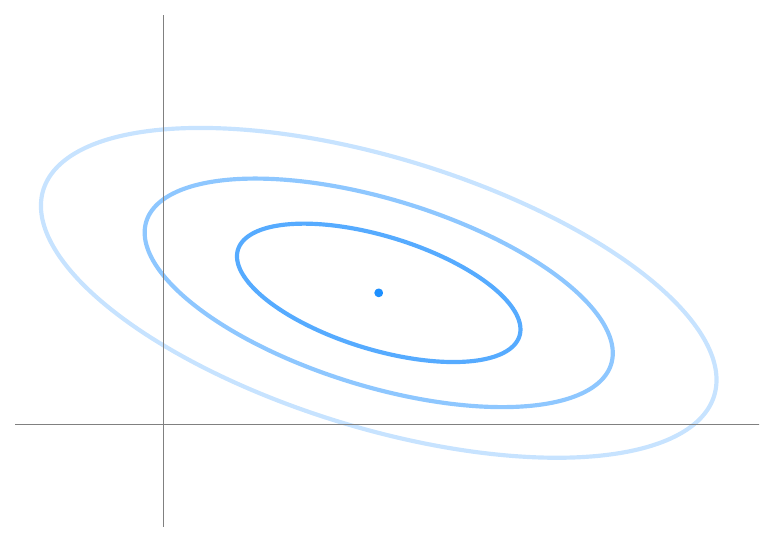} 
           } node [black, above] {\actiontxt{load}} (s5)
           
           (s5) edge [bend right, below, red] node [black, below=-0.2cm] {
           \includegraphics[width=1.0cm]{ 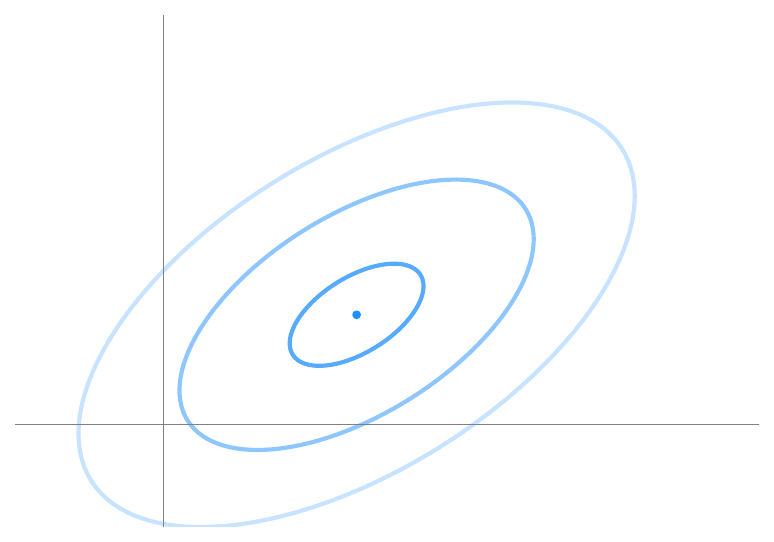} 
           } node [black, above] {\actiontxt{unload\_2}} (s2)

           (s3) edge [bend right, below, black] node [black, below=-0.2cm] {
           \includegraphics[width=1.0cm]{ 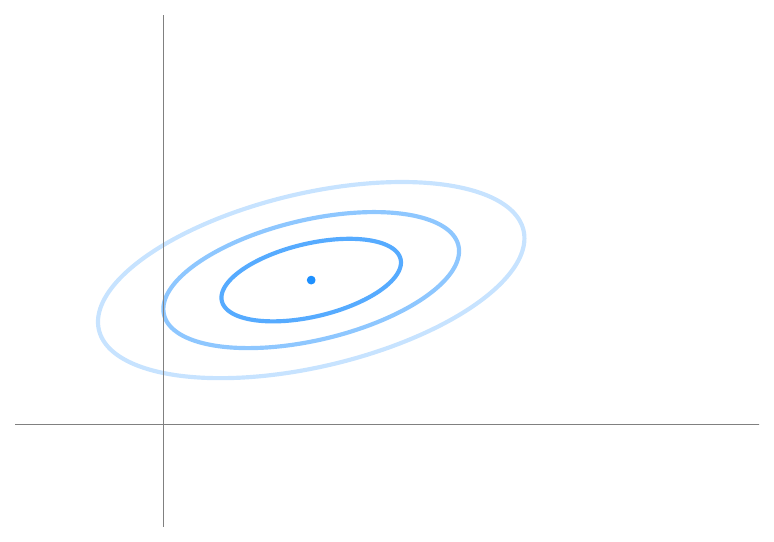} 
           } node [black, above] {\actiontxt{load}} (s5)
           
           (s5) edge [bend right, below, black] node [black, below=-0.2cm] {
           \includegraphics[width=1.0cm]{ 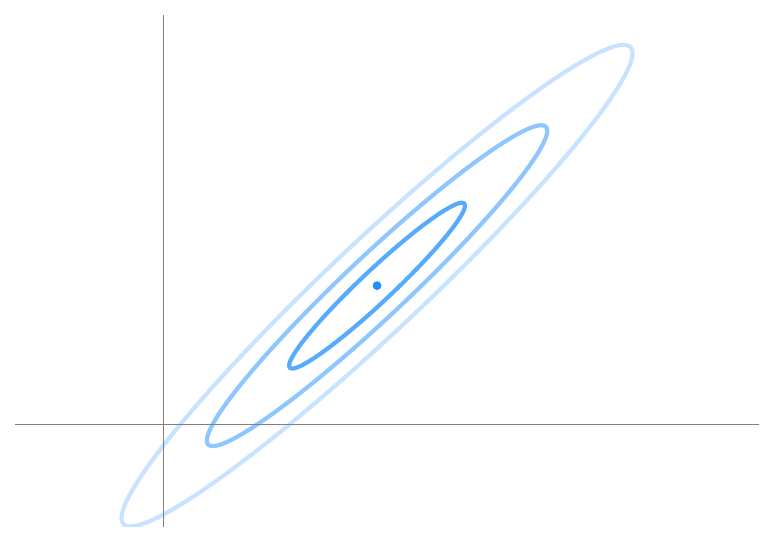} 
           } node [black, above] {\actiontxt{unload\_3}} (s3)

           (s4) edge [bend right, below, black] node [black, below=-0.2cm] {
           \includegraphics[width=1.0cm]{ 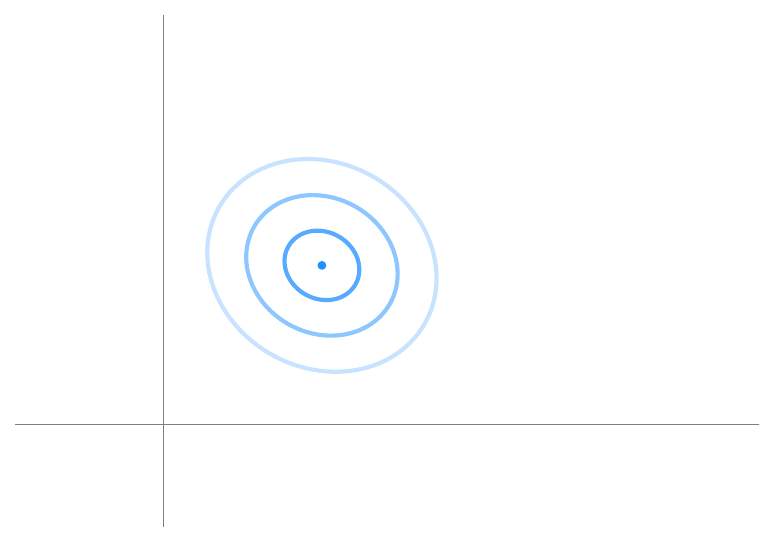} 
           } node [black, above] {\actiontxt{load}} (s5)
           
           (s5) edge [bend right, below, black] node [black, below=-0.2cm] {
           \includegraphics[width=1.0cm]{ 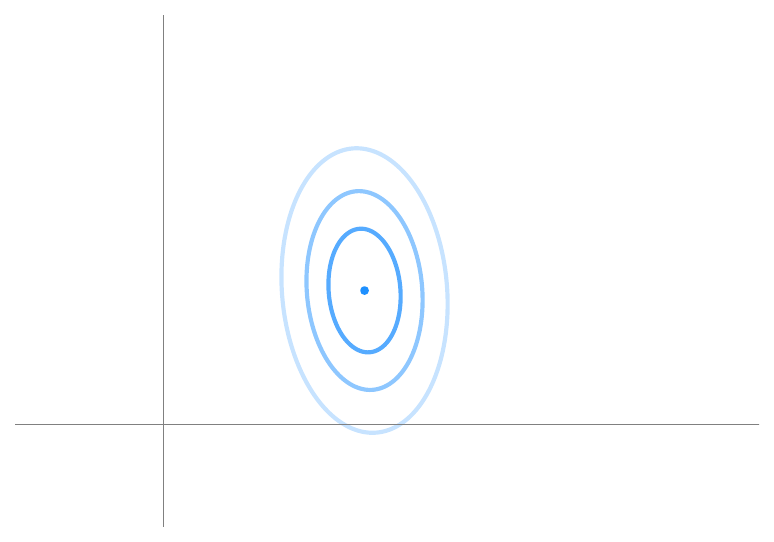} 
           } node [black, above] {\actiontxt{unload\_4}} (s4);

           \node[draw, align=center, below left=0.7cm and 0.5cm of s2, text width=2cm] (textbox) {$p(\cvec|s,a;\theta_{s,a})$};
           \coordinate (c_sa_dist) at ($(s2.south)!0.5!(s5.north) + (-0.7cm, -0.7cm)$);
           \draw[->, black, dashed] (textbox) -- (c_sa_dist);
    \end{tikzpicture}
    \end{adjustbox}
    \caption{The \DTS $T$ described in Example \ref{ex: dts} is shown, with a plan (red) that satisfies the task described in Example \ref{ex: formula}.}
    \label{fig: dts}
\vspace{-4 mm}
\end{figure}

\begin{example}
    Consider a simplified \DTS model in Fig. \ref{fig: dts} of the dishwashing scenario described in Fig. \ref{fig: scenario} with five states capturing the locations of each dish ($J_f$: jar on floor, $P_r$: pitcher on rack, etc.), and two actions \emph{\actiontxt{load}} and \emph{\actiontxt{unload}}. Each state observes either $\{dry\}$ or $\{wash\}$. Each action from a state takes time and incurs some risk when transporting the jar, modeled with a unique multivariate cost distribution for each state-action pair.
    \label{ex: dts}
\end{example}

When robot takes action $a\!\in\!A$ at state $s\!\in\!S$, it endures $N$ distinct types of costs $\cvec_{s,a}\!=\!(c_1, \ldots, c_N) \in \reals^N$, e.g., energy expenditure and time of execution. This cost vector can generally be stochastic, i.e., the cost value of executing $a$ at $s$ can differ at different instances due to, e.g., changes in the environment or use of randomized algorithms (sampling-based planners). 
Hence, $\cvec_{s,a}$ is a random variable whose distribution is given by $\mathcal{C}(s,a)$, i.e., $\cvec_{s,a} \sim \mathcal{C}(s,a)$.  
With an abuse of notation, we use $\cvec$ as the cost random variable for all state-action pairs, and hence $(\cvec\mid s,a) = \cvec_{s,a}$.

We treat $\mathcal{C}(s,a)$ as a conditional distribution of a probabilistic transition-cost generative model, parameterized by the set of parameters $\Theta$, i.e., 
{\allowdisplaybreaks
\abovedisplayskip = 3.5pt
\abovedisplayshortskip = 3.5pt
\belowdisplayskip = 3.5pt
\belowdisplayshortskip = 1.5pt
\begin{align}
\mathcal{C}(s,a) = p(\cvec \mid s, a; \Theta) \nonumber.
\end{align}
}For ease of presentation, we denote $(s,a)$ specific parameters by $\theta_{s,a} \subseteq \Theta$ such that $\Theta = \cup_{s \in S, a \in A} \theta_{s,a}$. 
We assume robot model $T$ is a Markov process. That is, 
$\cvec$ for each state-action pair is independent, i.e., $p(\cvec\!\mid\!s, a;\theta_{s,a})$ 
is independent from $p(\cvec\!\mid\!s', a';\theta_{s',a'})$ for all $(s,a)\neq (s',a') \in S\times A$. Further, we assume each $p(\cvec\!\mid\!s, a; \theta_{s,a})$ is a multivariate normal (MVN) distribution 
$\N(\mathbf{\mu}_{s,a}, \mathbf{\Sigma}_{s,a})$ paramerized by a non-negative mean $\mathbf{\mu}_{s,a} \in \mathbb{R}_{\geq 0}^N$ and covariance $\mathbf{\Sigma}_{s,a} \in \mathbb{R}^{N \times N}$, i.e., $\theta_{s,a}\!=\!\{\mathbf{\mu}_{s,a}, \mathbf{\Sigma}_{s,a}\}$.
Note that this assumption allows for correlation between types of costs. For instance, if the execution of a robotic action takes a long time, it is likely that it also uses more energy. 

In this work, we consider the realistic case where 
$\Theta$ 
are \emph{unknown}, i.e., both $\mean_{s,a}$ and $\cov_{s,a}$ are unknown for every $(s,a)$. 
One of the goals of this work is to learn these parameters
from data collected on $\cvec_{s,a}$ during execution.

Given a plan $\pi$
and its induced trajectory $\tau$
, the \textit{cumulative cost} of executing $\pi$ from state $s_0$, denoted $\ccvec_{s_0, \pi}$, is the sum of the state-action costs along $\tau$, i.e.,
$\ccvec_{s_0, \pi} = \sum_{k=0}^{m} \cvec_{s_k,a_k}.$
Since each $\cvec_{s_k,a_k}$ is a random variable,  $\ccvec_{s_0, \pi}$ is also a random variable, and its distribution is given 
by the convolution of the $\cvec_{s_k,a_k}$ distributions, i.e.,
\begin{align} \label{eq: ccost}
    p(\ccvec\!\mid\!s_0,\!\pi; \theta_{s_0,\pi})\!=\!p(\cvec \!\mid\!s_0, a_0; \Theta)\!\ast\!\ldots\!\ast \! p(\cvec\!\mid\!s_m, a_m;\Theta),
\end{align}
where $\theta_{s_0, \pi}$ is constructed by $\Theta$. Recall that $p(\cvec |s_k,a_k; \Theta)$ is assumed to be a MVN; 
$p(\ccvec \mid s_0, \pi; \theta_{s_0, \pi})$ 
is also a MVN. 
Since true 
$\Theta$ are unknown beforehand, we learn the parameters from experience. 
We denote the learned parameters by $\tilde \Theta$.

\subsection{Robotic Task} \label{sec: robotic_task}

The robot is given a complex task that can be completed in finite time.
To specify such tasks in a formal manner, we use \textit{Linear Temporal Logic over Finite Traces} (LTLf) 
\cite{de2013linear}, which combine propositional logic with temporal operators. 

\begin{definition}[LTLf Syntax] 
    A \emph{Linear Temporal Logic over Finite Traces} (LTLf) formula over $AP$ is recursively defined as
    \begin{align*}
        \phi := true \mid o \mid \neg\phi \mid \phi \land \phi \mid X\phi \mid \phi\, U \phi,
    \end{align*}
    where $o \in AP$, $\neg$ (negation) and $\land$ (conjunction) are Boolean operators, and $X$ (next) and $U$ (until) are temporal operators.
    \label{def:csltl}
\end{definition}
\noindent
The commonly-used temporal operators ``eventually" ($F$) and ``globally'' ($G$) can be defined as $F\phi \equiv true \, U \phi$ and $G\phi \equiv \neg F \neg \phi$, respectively.

The semantics of LTLf are defined over finite traces~\cite{de2013linear}.  That is,
an LTLf formula $\phi$ can be satisfied with a finite trace $w\in (2^{AP})^*$ \cite{de2013linear}, denoted $w \models \phi$. 
We define the language of $\phi$ to be the set of finite traces that satisfy $\phi$ for the first time, i.e., let $|w|$ be the length of $w$ and $w[k]$ for $1 \leq k \leq |w|$ be the prefix of $w$ with length $k$; the \textit{language} of $\phi$ is defined as
$$\mathcal{L}_\phi = \{w \in (2^{AP})^* \mid w \models \phi \land w[k] \not\models \phi \;\; \forall k <|w| \}.$$
We say plan $\pi \in A^*$ executed from $s$ satisfies $\phi$ if the resulting trace $L(\pi,s) \in \L_{\phi}$.
\begin{example}
    Following Example~\ref{ex: dts}, we specify the task $\phi = F(wash \land F(dry))$ stating ``the robot should first wash the dishes, and then dry the dishes.''
    \label{ex: formula}
\end{example}

We want the robot to execute task $\phi$ repeatedly.  Specifically,
we are interested in synthesizing plans that episodically repeat 
$\phi$. That is, once a plan is complete ($\phi$ is satisfied), a new plan has to be synthesized from the current state to satisfy $\phi$ again.  
We assume that such a repetition is always physically possible for the robot, i.e., the robot always ends in a state from which there exists a plan that satisfies $\phi$.  For instance, in the above example, after drying dishes, the robot always ends in a state from which it can start washing a new pile of dirty dishes.
Thus, we focus on the set of satisfying plans from every state, and refer to it as the $\phi$-satisfying plan set.
\begin{definition}[$\phi$-Satisfying Plan Set]
    Given \DTS robot model $T$, state $s \in S$, and LTLf task formula $\phi$, \emph{$\phi$-Satisfying Plan Set} $\Pi(s,\phi)$ is defined as 
    $$\Pi(s,\phi) = \{\pi \in \planset \mid L(\pi, s) \in \L_\phi \}.$$
    \label{def:macroaction_set}
\end{definition}

\subsection{Optimization Objectives} \label{sec: optimization_objectives}
At each planning (decision) instance from state $s$, we aim to select a plan $\pi^\star \in \Pi(s,\phi)$ that, in addition to completing task $\phi$, optimizes for the expected cumulative cost $\expec{\ccvec \mid s,\pi;\ \Theta}$.
Since $\ccvec_{s,\pi}$ is $N$-dimensional, it is in fact an $N$-objective optimization problem.
Moreover, the cost objectives may be competing, i.e., by optimizing one objective, another becomes sub-optimal. Hence, 
we want to optimize for the trade-offs between the objectives, which is known as Pareto optimal and best defined by the notion of vector dominance.  

\begin{definition}[Vector Dominance]
    Let $C,C' \in \reals^N$ be two vectors and $C(i)$ be the $i$-th element of $C$. Vector $C$ \emph{dominates} $C'$, denoted by $C \prec C'$, iff $C(i) \leq C'(i)$ for every $1 \leq i \leq N$ and there exists an $i$ such that $C(i) < C'(i)$.
\end{definition}

\begin{definition}[Pareto Optimal] \label{def:pf}
    For state $s \in S$ and LTLf formula $\phi$, plan $\pi \in \Pi(s,\phi)$ is called \emph{Pareto optimal} if 
    $$\nexists \pi' \in \Pi(s,\phi) \; \text{ \st } \;\; \expec{\ccvec \mid s,\pi'; \ \Theta} \prec \expec{\ccvec \mid s,\pi; \ \Theta}.$$
    The set of all Pareto optimal plans from $s$ is denoted by $\Pi^\star(s,\phi) \subseteq \Pi(s,\phi)$.
    Then, for each $\pi^\star \in \Pi^\star(s,\phi)$,
    $\expec{\ccvec \mid s,\pi^\star; \ \Theta}$ is called a \emph{Pareto point}, and the set of all Pareto points is called the \emph{Pareto front}, denoted $\Omega(s, \phi)$.  
    
\end{definition}

At every planning instance (episode), we desire to select a Pareto optimal plan $\pi^\star \in \Pi^*(s,\phi)$ which ensures that the expected cost is not dominated by another plan.  
However, since 
$\Theta$ are not available, it is impossible to compute for $\pi^\star$. Instead, we can use $\tilde \Theta$ to compute a learned set of $\phi$-satisfying Pareto optimal plans denoted $\tilde{\Pi}^\star(s,\phi) \subseteq \Pi(s,\phi)$. 
Similarly, let $\tilde \Omega(s,\phi)$  denote the set of expected cost vectors of each $\pi \in \tilde{\Pi}^\star(s,\phi)$. 
\subsection{User Preference} 
\label{sec: user_preference} 
There are possibly many Pareto optimal plans that perform 
$\phi$.  We are interested in picking a plan $\pi^\star_{\pref}$ that achieves the user's \emph{most-preferred} Pareto point in every instance. Given that our goal is autonomous operation, we require this preference to be provided as an input before deployment.
This poses a challenge because the Pareto front is unknown \textit{a priori}, and the user cannot know the exact Pareto point they prefer.  
Instead, as suggested in recent literature \cite{lanillos2021, wakayama2023}, the user can provide a probability density function (pdf) that serves as an expressive way of capturing preferred observed outcomes. We extend this notion to the multi-objective domain, and leverage the expressive pdf preference as a tie-breaker between optimal trade-off candidates.

We specifically focus on user preference as a normal distribution $p_{\pref}(\ccvec) = \N(\mean_{\pref},\cov_{\pref})$ over the objectives since it can be fully defined by only specifying a mean $\mu_{\pref} \in \reals^N$ and covariance $\cov_{\pref} \in \reals^{N\times N}$, respectively. Thus, by providing $\mean_\pref$ and $\cov_\pref$, the user can not only express a desired trade-off region in the objective space, but also specify the willingness to explore alternative trade-offs (Pareto points) that allow the robot to learn more quickly, offering means to address the \textit{exploration-exploitation trade-off} problem.

\begin{example}
    Suppose the dishwashing robot described in Fig.~\ref{fig: scenario} is being deployed in a restaurant. If the restaurant is going to be busy, the user may define a preference distribution with mean in a low-time high-risk part of the objective space, whereas if the restaurant will be slow, the user may opt for a high-time, low-risk alternative, both of which can be informed by past observations of how humans wash the dishes. If the user is interested in learning other optimal trade-offs, or is uncertain how accurately they have estimated the mean, they can expand and orient the covariance around regions they would like the robot to explore information rich opportunities.
    \label{ex: preference}
\end{example}

To meet user's preference, we seek a Pareto point that is not ``suprising.'' The information theoretic definition of {\it surprise} is the negative log of model evidence \cite{cover2006}. 
Similarly, we define the surprise of an observation obtained by following a plan $\pi$ as:
\begin{align} \label{eq: surprise}
    \mathrm{Surprise}(\pi)\!=\!-\log p(\ccvec \mid \pi).
\end{align}
This quantity can be derived by marginalizing the following generative model:
\edt{
\begin{align} \label{eq: surprise_joint}
    (\ref{eq: surprise}) &= - \log \int_{s} \int_{\tilde{\theta}_{s,\pi}} p(s, \tilde{\theta}_{s,\pi}, \ccvec \mid \pi)  d\tilde{\theta}_{s,\pi} ds, \nonumber \\
    &= - \log \int_{s} \int_{\tilde{\theta}_{s,\pi}} p(s, \tilde{\theta}_{s,\pi} \mid \ccvec, \pi)p(\ccvec \mid \pi) d\tilde{\theta}_{s,\pi} ds. 
\end{align}}
In active inference (AIF), one can inject a bias towards seeing certain outcomes by substituting the predictive outcome distribution $p(\ccvec \mid \pi)$ with the prior preference $p_{\pref}(\ccvec)$ (a.k.a. ``evolutionary prior") that is independent of any plan ~\cite{smith2022,friston2022}. We refer to the biased surprise as \textit{surprise w.r.t.} $p_\pref$:
\edt{
\begin{align} \label{eq: biased_surprise}
    \mathrm{Surprise}(\pi,p_\pref)\!=\!-\! \log \int_{s}\!\int_{\tilde{\theta}_{s,\pi}}\! p(s,\!\tilde{\theta}_{s,\pi} | \ccvec,\!\pi)p_{\pref}(\ccvec) d\tilde{\theta}_{s,\pi} ds.
\end{align}}
In essence, under AIF, an agent acts to see desired observations. 
An overview of AIF is provided in Sec. \ref{sec: active_inference}. 

\subsection{Problem Statement} \label{sec: problem_statement}

The problem we consider is as follows.
\begin{problem} \label{problem}
    Consider a \DTS robot model $T$ with unknown cost distributions, an LTLf task specification $\phi$ that the robot is to repeatedly complete, and a user preference distribution $p_{\pref}(\ccvec)$ over $N$ cost objectives.  Let $K \in \naturals$ denote the number of times (instances) the robot achieves task $\phi$, and $s_K \in S$ be the start state of the robot in the $K$-th episode.
    Then, for each $K$, synthesize a 
    $\phi$-satisfying plan $\pi_K \in \Pi(s_K,\phi)$ such that 
    \begin{enumerate}
        \item[(i)] $\pi_K$ minimizes $\mathrm{Surprise}(\pi, p_\pref)$ in \eqref{eq: biased_surprise},
        \item[(ii)] as $K\rightarrow \infty$, $\pi_K$ becomes Pareto optimal, i.e., $\pi_K \in \Pi^\star(s_K,\phi)$, and
        \item[(iii)] as $K\rightarrow \infty$, the resulting Pareto front $\tilde{\Omega}(s_K, \phi)$ converges to the true Pareto front $\Omega(s_K, \phi)$.
    \end{enumerate}
\end{problem}

Note that, in Problem \ref{problem},  satisfaction of $\phi$ (qualitative objective) at each instance is a hard constraint, and the optimization (quantitative) objectives in (i)-(iii) are soft constraints.  Hence, by incorporating safety requirements in $\phi$, safety satisfaction is guaranteed throughout the process.
Furthermore, from the perspective of multi-objective decision making, Problem \ref{problem} is particularly challenging since adhering to a user's preference and learning the Pareto front are arguably perpendicular efforts. 
In our approach, we address 
this 
challenge by using free energy minimization to naturally balance exploitation and exploration over the Pareto front.

%


\section{Approach} \label{sec: approach}

\begin{figure*}[t]
    \centering
    \includegraphics[width=\linewidth]{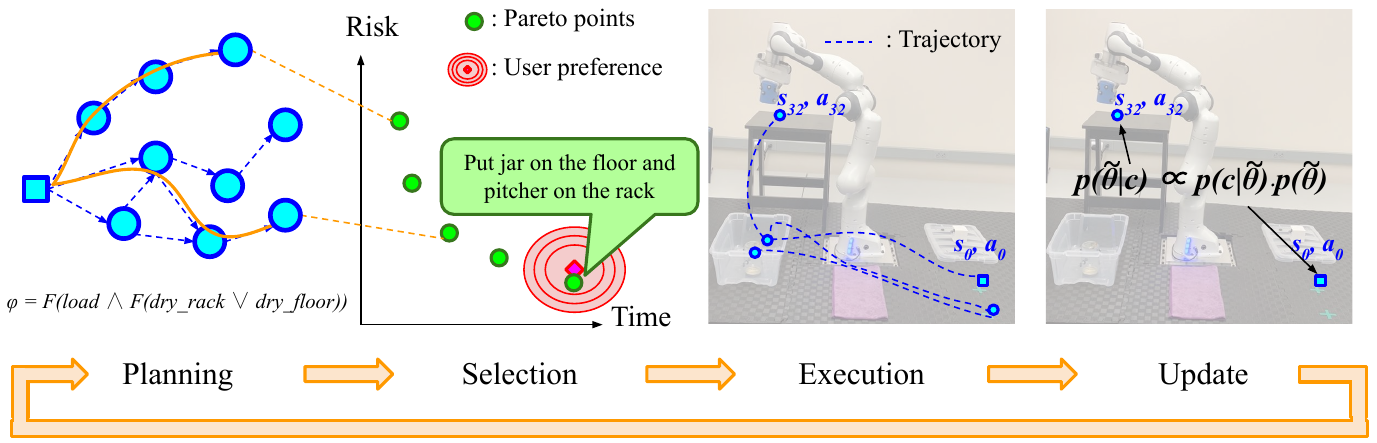}
    \caption{Proposed multi-objective safe reinforcement learning architecture: As the four phases involving 1) Planning, 2) Selection, 3) Execution, and 4) Update are repeatedly executed, the robot gradually learns the state-action cost parameters and selects the user preferred optimal task plan.}
    \label{fig: proposed_approach}
\end{figure*}

Our approach to Problem~\ref{problem} is an iterative MORL framework with four phases: 1) \textit{planning} Pareto optimal plans, 2) \textit{selection} of the preferred Pareto optimal plan, 3) \textit{execution} of the selected plan, and 4) \textit{update} to the executed state-action cost parameters, as illustrated in Fig. \ref{fig: proposed_approach}. 
\textit{Planning}, \textit{selection}, and \textit{update} all are dependent on the learning status of the robot. 
Hence, we start by formalizing our approach to modeling $\tilde \Theta$.

\subsection{Learning the Transition Cost Model} \label{sec: cost_estimate_model}
Recall that the cost $\cvec_{s,a}$ 
of each $(s,a)$ 
is distributed as $p(\cvec \mid s,a; \theta_{s,a}) = {\cal N}(\mu_{s,a}, \Sigma_{s,a})$.
However, $\theta_{s,a}$ are unknown and must be learned using experience. 
We take a Bayesian approach to learning $\theta_{s,a}$ by modeling with random variable $\tilde{\theta}_{s,a}$. 
Specifically, we assume that the robot collects a realized sample $\hat{\cvec}$ drawn from $p(\cvec|s,a;\theta_{s,a})$ each time $a$ is executed from $s$. Let $D_{s,a}\!=\!\{\hat{\cvec}_1, \ldots \hat{\cvec}_l \}$ be the set of $l$ realizations of $\cvec_{s,a}$. We maintain a belief over $\tilde{\theta}_{s,a}$ using the Normal Inverse-Wishart (NIW) distribution, i.e., 
$$p(\tilde{\theta}_{s,a})\!=\!NIW_{s,a}(\lambda_0, \kappa_0, \Lambda_0, \nu_0).$$ 
NIW is chosen because of the property of conjugacy for estimating unknown mean and covariance of a MVN distribution, i.e., the posterior is also \edt{represented analytically as} a NIW \cite{murphy2007conjugate} $$p(\tilde{\theta}_{s,a}\!\mid\!D_{s,a})\!=\!NIW_{s,a}(\lambda, \kappa, \Lambda, \nu).$$
\edt{Note that some works assume the covariance is known \textit{a priori} \cite{reverdy2014modeling}, limiting the fidelity of the model. NIW strikes a balance between an over simplified parametric model (known covariance) and a sample inefficient non-parametric model \cite{chen2017tutorial}.}
We detail the the update procedure in Sec. \ref{sec: update}. 
\edt{If the user has an informed guess about the mean and covariance of a certain state-action distribution, they can set $\lambda_0$ and $\Lambda_0$ respectively, reducing the}
number of decision making instances required to learn $\tilde{\Theta}$.


\subsection{Planning}
At the start of each planning (decision-making) instance, we employ a multi-objective task planner to compute the set of approximate Pareto-optimal plans $\tilde{\Pi}^\star(\scurr, \phi)$ described in Def. \ref{def:pf}.
To this end, we formulate a \textit{correct-by-construction} multi-objective shortest-path graph search problem, similar to \cite{amorese2023}, which can be solved with the following procedure.

\subsubsection{Product Graph Construction}
An LTLf formula can be translated into a deterministic finite automaton (DFA) \cite{de2013linear}, a finite state-machine 
that precisely captures the language of $\phi$.

\begin{definition} [Deterministic Finite Automaton]
    A \emph{DFA} constructed from LTLf formula $\phi$ is a tuple $\Q = (\Gamma, \gamma_0, W, \delta_{\Q}, \mathcal{F})$ where $\Gamma$ is a set of states, $\gamma_0 \in \Gamma$ is the initial state, $W = 2^{AP}$ is the alphabet, $\delta_{\Q} : \Gamma \times W \mapsto \Gamma$ is a deterministic transition function, and $\mathcal{F}$ is a set of accepting states.
\end{definition}

A trace $w = w_0 \ldots w_{m-1} \in (2^{AP})^*$ induces a finite run $\tau^{\Q} = \gamma_0 \gamma_1 \ldots \gamma_m$ on ${\Q}$, where $\gamma_{k+1} = \delta_{\Q}(\gamma_k, w_k)$. Run $\tau^{\Q}$ is called accepting if $\gamma_m \in \mathcal{F}$.  If $\tau^{\Q}$ is accepting, then trace $w$ is accepted by ${\Q}$ and satisfies $\phi$, i.e., $w \models \phi$.

%
%


Using $T$ to capture the physical capability of the robotic system, along with ${\Q}$ to capture the temporal attributes of $\phi$, we can construct a product automaton that intersects the restrictions of both $T$ and ${\Q}$.

\begin{definition} [Product Automaton]
    Given a \DTS $T$, current state $s_K \in S$, and DFA ${\Q}$, a \emph{Product Automaton} is a tuple $\mathcal{P} = (P, \rho_K, A, \delta_P, \mathbf{v}, \mathcal{F}_P)$, where 
    \begin{itemize}
        \item $P = S \times \Gamma$ is a set of states,
        \item $\rho_K = (s_K, \gamma_0')$, where $\gamma_0' = \delta_\Q(\gamma_0, L(s_K))$, is the starting state in the current instance,
        \item $A$ is the same action set in $T$,
        \item $\delta_P : P \times A \mapsto P$ is a transition function, where for states $\rho = (s, \gamma)$ and $\rho' = (s', \gamma')$ and action $a \in A$, transition $\rho' = \delta_P(\rho, a)$ exists if $s' = \delta_T(s,a)$ and $\gamma' = \delta_{\Q}(\gamma, L(s'))$,
        \item $\mathbf{v} : P \times A \mapsto \mathbb{R}_{\geq 0}^N$, is a transition cost-vector function, and
        \item $\mathcal{F}_P = S \times \mathcal{F}$ is a set of accepting states.
    \end{itemize}
\end{definition}
\noindent
Note that unlike stochastic cost function $\C$ in $T$, cost-vector function $\mathbf{v}(\rho, a) \in \reals_{\geq 0}^N$ is not a distribution.  We elaborate on $\mathbf{v}$ further below.

Similar to $T$, a plan $\pi$ induces a run $\tau^P = \rho_0 \ldots \rho_{m+1}$ on $\mathcal{P}$ where $\rho_{k+1} = \delta_P(\rho_k, a_k)$. Run $\tau^P$ is \textit{accepting} iff $\rho_{m+1} \in \mathcal{F}_P$. 
Therefore, by construction, the set of $\phi$-satisfying plans $\Pi(\scurr, \phi)$ is equal to the set of all plans that induce paths on $\mathcal{P}$ from $\rho_K$ to a $\rho \in \mathcal{F}_P$.

\subsubsection{Transition Cost-Vector Function}
We aim to define cost-vector function $\mathbf{v}$ and accordingly formulate a multi-objective graph search problem on $\mathcal{P}$ that enables the computation of the set of Pareto optimal $\phi$-satisfying plans $\Pi^*(s_K, \phi)$.
Consider current state $\rho_K = (s_K,\gamma_0)$, plan $\pi \in \Pi(s_K, \phi)$, and induced path $\tau(\rho_K,\pi) = \rho_{K,0} \ldots \rho_{K,m+1}$.
Ideally, for every $\rho = (s,\gamma) \in P$ and $a \in A$, we want $\mathbf{v}((s,\gamma),a) = \expec{\cvec | s, a; \Theta}$ 
because the total cost of induced path $\tau^P(\rho_K,\pi)$ becomes
\begin{align*}
    \sum_{i = 0}^{m+1} \mathbf{v}(\rho_{K,i},\pi_i) &= \sum_{i = 0}^{m+1} \expec{\cvec | s_{K,i}, \pi_i; \Theta} 
    = \expec{\ccvec | s_K, \pi; \theta_{\scurr, \pi}}.
\end{align*}
Therefore, by assigning $\mathbf{v}(\rho,a) = \expec{\cvec | s, a; \Theta}$, the multi-objective Pareto front of the graph search problem on $P$ is exactly equal to $\Omega(\scurr, \phi)$ (see Def.~\ref{def:pf}). Then, existing algorithms such as a multi-objective variant of Dijkstra's algorithm or $A^*$ \cite{mandow2008multiobjective} can be employed to compute $\Omega$.
\edt{Note that these multi-objective graph search algorithms are exponential in the worst case, however the average branching factor for a $N$-dimensional vector belonging to a set of size $\mathcal{G}$, i.e. the set of all unique cost vector edge weights, is reduced to $O((\log |\mathcal{G}|)^{N-1})$, with certain conditions on the ordering \cite{mandow2008multiobjective}.}

However, $\Theta$ is unknown. A naive approach simply uses learned estimates $\expec{\tilde \Theta}$ for $\mathbf{v}$ instead, but that suffers from biasing the solution towards taking transitions that have low estimate cost, regardless of how accurate $\tilde \Theta$ is. To address this, we interweave exploration into the graph search problem, as detailed below. 

\subsubsection{Pareto-Regret} \label{sec: regret}
The efficacy of learning can be analyzed with the notion of cumulative ``Pareto-regret'', quantifying how sub-optimal each plan is with respect to the true Pareto front $\Omega(\scurr, \phi)$. Formally, given 
$\pi$ with true mean cumulative cost $\mu_{\scurr, \pi} = \expec{\ccvec | \scurr, \pi; \Theta}$, the Pareto regret for a given instance is defined as 
\begin{align} \label{eq:pareto-regret}
    r(&\scurr, \pi) = \nonumber \\
    &\min_{\epsilon} \{\mu_{\scurr, \pi}\!-\!\epsilon \mathbf{1} \mid \mu^\star \not\prec \mu_{\scurr, \pi}\!-\!\epsilon \mathbf{1} \;\; \forall \mu^\star \in \Omega(\scurr, \phi) \},
\end{align}
where $\mathbf{1}$ is a vector of ones with appropriate dimension.
Note that if $\pi \in \Pi^\star(\scurr, \phi)$, then $r(\scurr, \pi) = 0$. The cumulative regret up to instance $K$ is simply the summed regret for every instance, i.e. $R = \sum_{i = 0}^{K} r(s_i, \pi_i)$. To minimize cumulative regret, we augment the transition cost-vector with an exploration strategy.

\subsubsection{Pareto Cost-LCB}
We adapt the well established Pareto-UCB1 \cite{drugan2013, auer2002} strategy for computing an estimate Pareto front that balances the current best estimate cost model (exploitation) with a bonus reduction in cost (exploration).
The learned mean transition-cost is calculated using the estimated parameters $\expec{\mathbf{\tilde{\theta}}_{s,a}}$. 
To maintain optimality guarantees of the aforementioned graph search method, we rectify each element of $\mathbf{v}$ to be non-negative. The cost-lower confidence bound (LCB) cost vector $\mathbf{v}(\rho,a)$ is computed element-wise as:
\begin{align*}
    v_i(\rho,a)\!=\!\max \left\{0, \; \expec{\cvec | s,a; \expec{\tilde{\theta}_{s,a}}}_i - \alpha  \sqrt{{\log(k_g)}/{n(s,a)}}\right\} 
\end{align*}
where $\alpha$ is a ``confidence'' hyperparameter, $k_g$ is the current global time step across all decision instances, and $n(s,a)$ is the number of times action $a$ has been executed from state $s$. 

Note that many multi-objective multi-armed bandit confidence bound approaches provide theoretical regret bounds that are logarithmic with respect to the number of decision instances. It is not straight forward to extend these regret bounds to this framework due to the exponential nature of the planning space $\Pi(\scurr, \phi)$ which has size $O(|A|^{|P|})$. Hence, we leave theoretical regret bounds to future research.

Using the techniques described above, we can now compute the approximate set of Pareto optimal plans $\tilde{\Pi}^\star(s_K, \phi)$ while accounting for exploration of the environment. The exploration in \textit{planning} aids in the effort in addressing goals (ii) and (iii) of Problem~\ref{problem}, however, to uphold proper learning of the Pareto front (iii), the selection must also explore among trade-offs.

%



\subsection{Pareto Point Selection} \label{sec: selection} 
\begin{figure}[t]
    \begin{adjustbox}{scale=0.7}
    \centering
    \begin{tikzpicture}[shorten >=1pt,node distance=6cm,>=stealth',thick, auto,
                        every state/.style={fill,very thick,black!20,text=black,\shadowString},
                        robot/.style = {fill,very thick,black!20,rounded corners, text=black, shape=rectangle, minimum height=1cm,minimum width=1cm, blur shadow},
                        accepting/.style ={blue!50!black!50,text=white,accepting by double},
                        initial/.style ={red!80!black!40,text=black,initial by arrow, initial left}, initial text=$ $]
                        
        \node[robot] (s5) {$J_d, P_d$};
        \node[robot] (s1) [below left=0.3cm and 2.5cm of s5] {$J_f, P_f$};
        \node[robot] (s2) [above left=2.1cm and 1.5cm of s5] {$J_r, P_f$};
        \node[robot] (s3) [above right=2.1cm and 1.5cm of s5] {$J_r, P_r$};
        \node[robot] (s4) [below right=0.3cm and 2.5cm of s5] {$J_f, P_r$};

        \path[->]
           (s1) edge [bend left, below, red] node [black, below=-0.2cm] {
           \includegraphics[width=1.0cm]{ 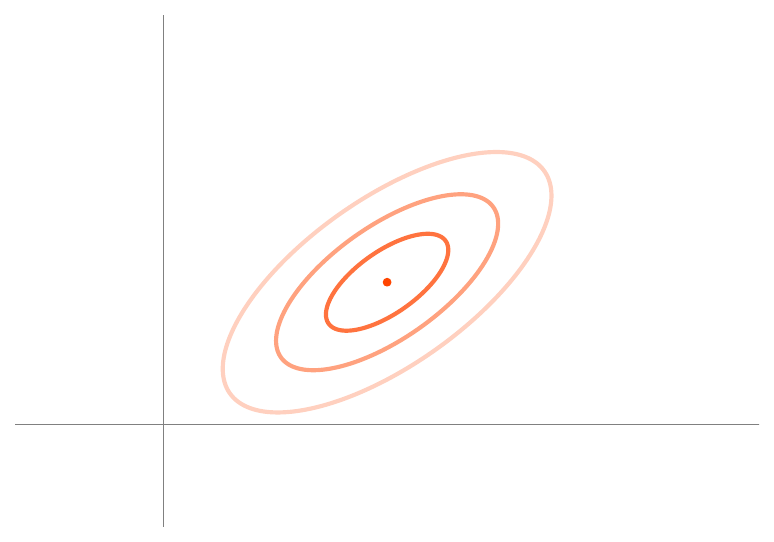} 
           } node [black, above] {$\pi_1 = $\actiontxt{load}, \actiontxt{unload\_2}} (s2)
           
           (s2) edge [bend left, below, black] node [black, below=-0.2cm] {
           \includegraphics[width=1.0cm]{ 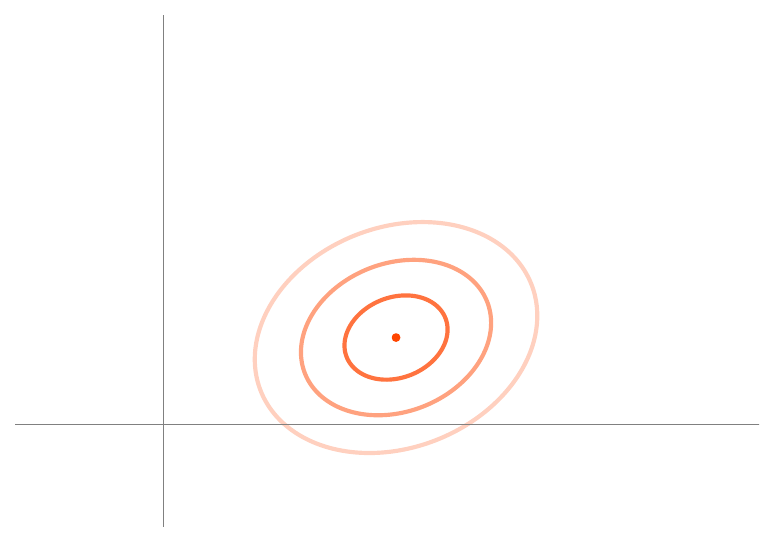} 
           } node [black, above] {$\pi_2$} (s1)

           (s2) edge [bend left, below, black] node [black, below=-0.2cm] {
           \includegraphics[width=1.0cm]{ 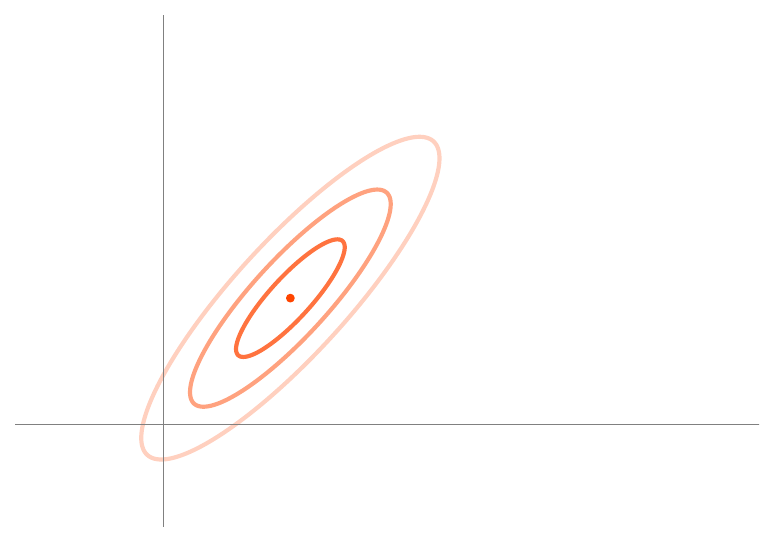} 
           } node [black, above] {$\pi_3$} (s3)
           
           (s3) edge [bend left, below, black] node [black, below=-0.2cm] {
           \includegraphics[width=1.0cm]{ 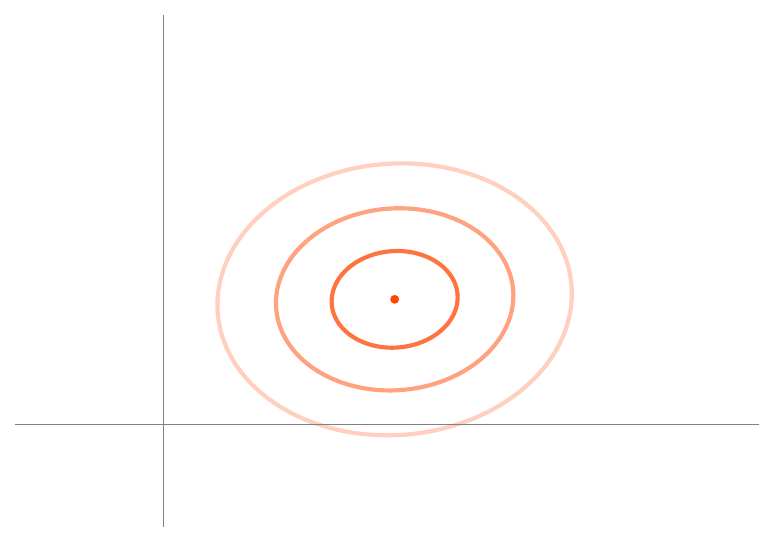} 
           } node [black, above] {$\pi_4$} (s2)

           (s3) edge [bend left, below] node [black, below=-0.2cm] {
           \includegraphics[width=1.0cm]{ 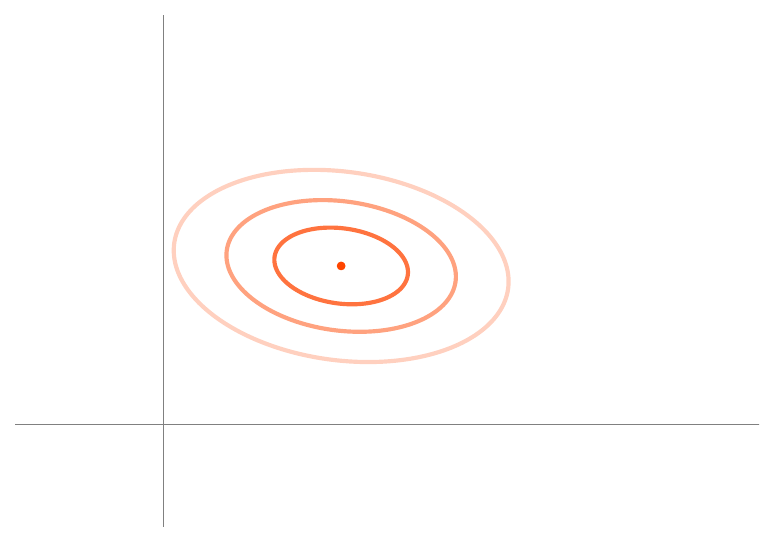} 
           } node [black, above] {$\pi_5$} (s4)
           
           (s4) edge [bend left, below, black] node [black, below=-0.2cm] {
           \includegraphics[width=1.0cm]{ 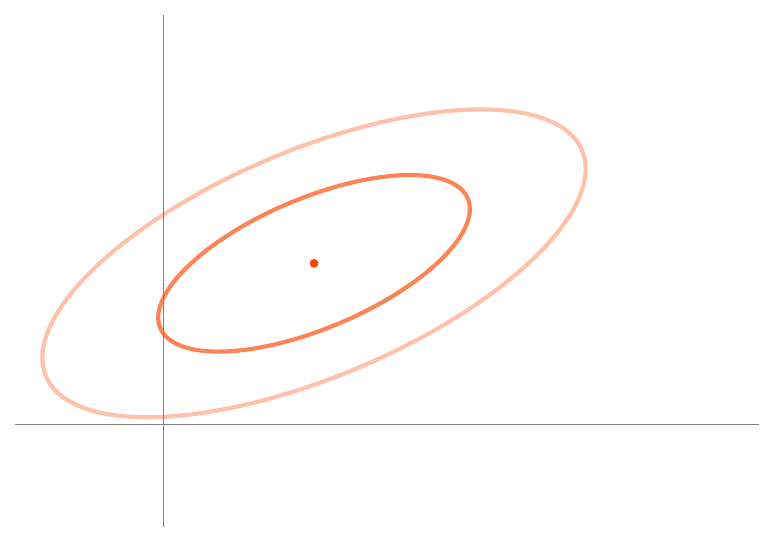} 
           } node [black, above] {$\pi_6$} (s3)

           (s5) edge [below, black] node [black, below=-0.2cm] {
           \includegraphics[width=1.0cm]{ 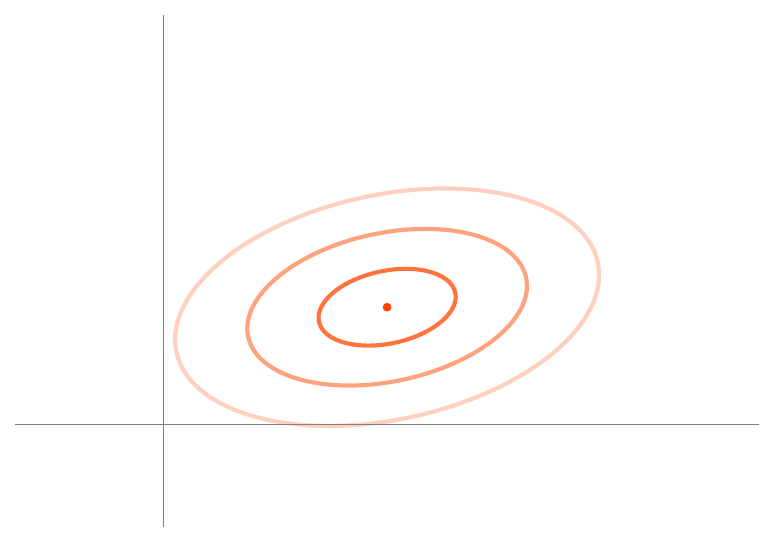} 
           } node [black, above] {$\pi_7$} (s1)
           
           (s5) edge [below, black] node [black, below=-0.2cm] {
           \includegraphics[width=1.0cm]{ 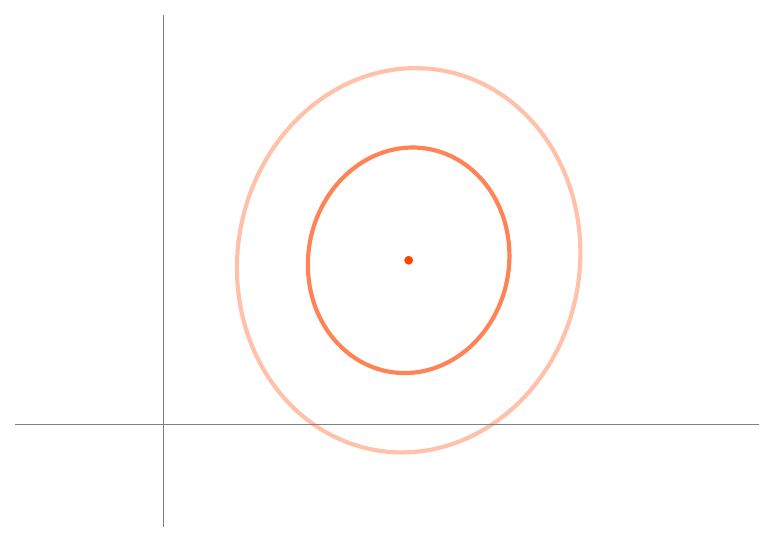} 
           } node [black, above] {$\pi_8$} (s2)
           
           (s5) edge [below, black] node [black, below=-0.2cm] {
           \includegraphics[width=1.0cm]{ 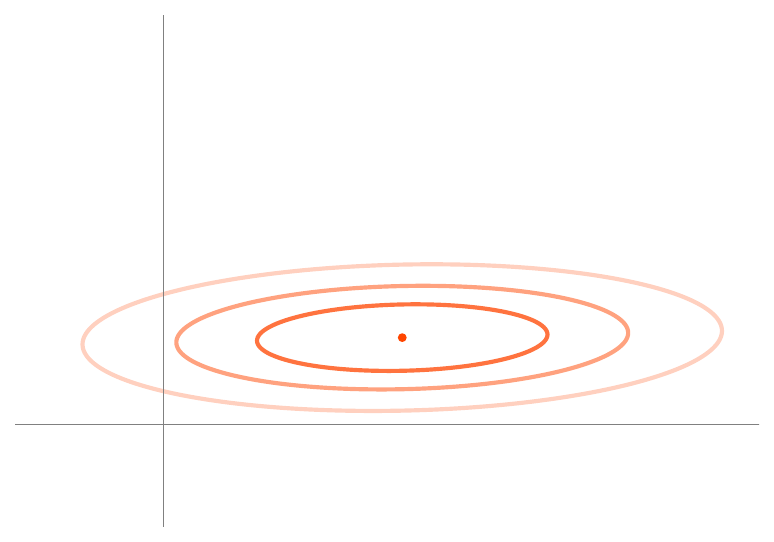} 
           } node [black, above] {$\pi_9$} (s3)
           
           (s5) edge [below, black] node [black, below=-0.2cm] {
           \includegraphics[width=1.0cm]{ 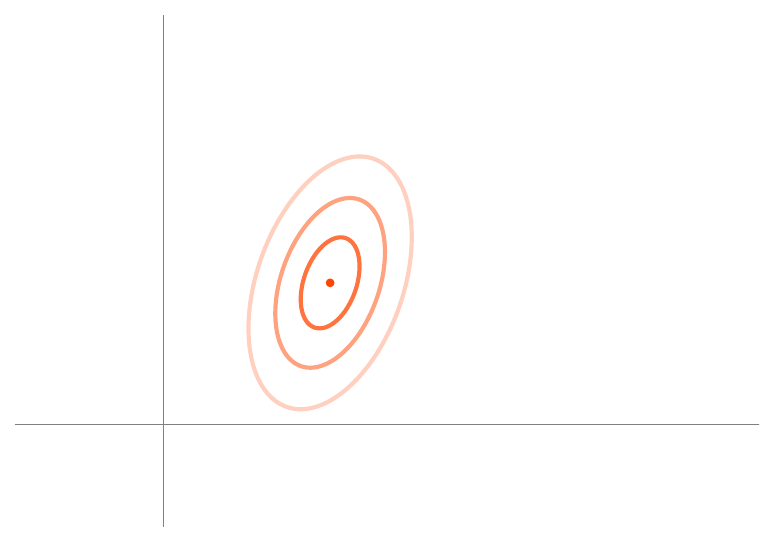} 
           } node [black, above] {$\pi_{10}$} (s4);
           
           \node[draw, align=center, left=0.2cm and 0.5cm of s1, text width=2cm] (textbox) {$p(\ccvec|s,\pi;\theta_{s,\pi})$};
           \coordinate (c_plan_dist) at ($(s1.south)!0.5!(s2.north) + (-0.9cm, -0.2cm)$);
           \draw[->, black, dashed] (textbox) -- (c_plan_dist);

    \end{tikzpicture}
    \end{adjustbox}
    \caption{By treating plans $\pi$ as macro-actions, the \textit{selection} decision making problem becomes sequential. The example plan highlighted in Fig. \ref{fig: dts} is embedded into $\pi_1$, as shown.}
    \label{fig: tedts}
\vspace{-4 mm}
\end{figure}
For the \textit{selection} problem, one can think of candidate plans $\pi \in \tilde{\Pi}^\star(s_K, \phi)$
computed by the planner as abstract macro-actions, such that high-level decision making can be done sequentially for each instance. 
Fig. \ref{fig: tedts} provides a visual representation of the abstracted selection problem with respect to the evolution of the robotic system over each instance.

As described in Sec. \ref{sec: user_preference}, ideally, the agent should select an action minimizing the surprise of outcomes. However, since it is analytically intractable to marginalize the joint distribution to derive the surprise, its upper bound, i.e., free energy, is minimized. Yet, the outcomes $\ccvec$ cannot be observed until a plan $\pi$ is actually executed, thus, the agents end up minimizing the so-called expected \edt{variational} free energy (EFE) in active inference based decision-making. In this Pareto point selection problem, the EFE is described as follows (see Appendix \ref{appendix: efe_pareto_derivation} for detailed derivation).
\begin{align} \label{eq: efe}
    &\mbox{EFE}(\scurr, \pi^\star, p_{\pref})\!=\!-\mathbb{E}_{q(\ccvec |\pi^\star)} \Big[ \log p_{\pref}(\ccvec) \Big] \nonumber \\
    &\!-\!\mathbb{E}_{q(\ccvec|\pi^\star)}\!\Big[\kld\!\Big(\!q(\scurr|\ccvec,\pi^\star) || q(\scurr|\pi^\star)\!\Big)\!\Big] \nonumber \\ 
    &\!-\!\mathbb{E}_{q(\ccvec|\pi^\star)}\!\Big[\kld\!\Big(\!q(\tilde{\theta}_{\scurr, \pi^\star}|\ccvec,\!\scurr,\!\pi^\star) || q(\tilde{\theta}_{\scurr, \pi^\star}|\scurr,\!\pi^\star)\!\Big)\!\Big],
\end{align}
where $q(\cdot)$ represents a proposal distribution. 
\edt{EFE is a variational quantity \cite{blei2017variational}, yielding freedom in the choice of proposal distribution $q(\tilde{\theta}_{\scurr, \pi} |\scurr,\!\pi^\star)$. As $q(\tilde{\theta}_{\scurr, \pi} |\scurr,\!\pi^\star)$ more closely resembles $p(\tilde{\theta}_{\scurr, \pi} | \ccvec, \scurr,\!\pi^\star)$, the upper bound on (expected) surprise decreases.} The first term of EFE in (\ref{eq: efe}) represents how much the predicted cost distribution $q(\ccvec|\pi^\star)$ aligns with 
$p_{\pref}(\ccvec)$ (i.e. exploitation), and the second and third terms represent how much the uncertainties of 
$\scurr$ and 
$\tilde{\theta}_{\scurr, \pi^\star}$ can be reduced by following 
$\pi^\star$ and measuring 
$\ccvec$ (i.e. exploration).
Since the user does not know the true Pareto-optimal region \textit{a priori}, EFE cannot be used to scalarize any plan in $\Pi$, as it may often be sub-optimal. Therefore, as described in Sec. \ref{sec: multi-objective_optimization}, the preferred plan $\pi_{\pref}\!=\!\argmin_{\pi \in \Pi}\mbox{EFE}(\pi, p_{\pref})$ is not guaranteed to also be an element in $\Pi^\star$ and vice versa. Consequently, in order to ensure that the user's preference does not bias the system away from optimal behavior, this preference is only expressed over Pareto-optimal candidates.
So, $\Pi^\star(\scurr, \phi)$ is scalarized via EFE in order to adhere to $p_{\pref}$ and reduce uncertainties in the belief over $\tilde \theta_{\scurr, \pi^\star}$.
Since $T$ models deterministic transitions, the second term of (\ref{eq: efe}) can be ignored. Expanding the third term of (\ref{eq: efe}) yields
\begin{align} \label{eq: efe_entropy}
    &\mbox{EFE}(\scurr, \pi^\star, p_{\pref}) = -\mathbb{E}_{q(\ccvec |\pi^\star)}\!\Big[\!\log p_{\pref}(\ccvec)\!\Big] \nonumber \\
    &-\entropy{q(\tilde{\theta}_{\scurr, \pi} |\scurr,\!\pi^\star)} 
     \!+\!\mathbb{E}_{q(\ccvec |\pi^\star)} \Big[ \entropy{q(\tilde{\theta}_{\scurr, \pi^\star} | \ccvec,\!\scurr,\!\pi^\star)} \Big],
\end{align}
where $\entropy{\cdot}$ represents the entropy of a pdf. Then, the preferred plan is  
\begin{equation} \label{eq:argminefe}
    \pi^\star_{\pref} = \argmin_{\pi^\star\in \Pi^\star(s_K)}\mbox{EFE}(s_K, \pi^\star, p_{\pref}).
\end{equation}

Due to the Markov property of \DTS and \eqref{eq: ccost},
$\mathbf{\mu}_{\scurr, \pi^\star}$ and 
$\mathbf{\Sigma}_{\scurr, \pi^\star}$ of the convolved distribution $p(\ccvec | \scurr, \pi^\star; \theta_{s_{K},\pi^\star})$ are parameterized by 
$\sum_{k}\mu_{s_k,a_k}$ and 
$\sum_{k}\mathbf{\Sigma}_{s_k,a_k}$ of each cost distribution respectively. According to the estimate transition cost model, 
$\tilde{\theta}_{s,a}\!=\!\{\tilde{\mu}_{s,a}, \tilde{\mathbf{\Sigma}}_{s,a} \}$ are random variables themselves; therefore, 
\begin{equation} \label{eq:intractable}
    p(\tilde{\theta}_{\scurr, \pi^\star} |\scurr, \pi^\star) = p(\tilde{\theta}_{s_0,a_0}) \ast \ldots \ast p(\tilde{\theta}_{s_m,a_m}).
\end{equation}
As described in Section \ref{sec: cost_estimate_model}, each $p(\tilde{\theta}_{s,a})$ 
is a unique NIW, which makes (\ref{eq:intractable}) analytically intractable. 
To enable computation for $\pi_\pref^\star$ in \eqref{eq:argminefe}, \edt{we leverage both the functional freedom in choosing proposal distributions $q(\cdot)$ as well as three} statistical approximations of \eqref{eq: efe_entropy}: certainty equivalence of the predicted observation distributions \cite{mania2019certainty}, central limit theorem (CLT) \cite{kwak2017central}, \edt{and Monte Carlo sampling \cite{bishop2006}}.


\subsubsection{Approximating the First Term of (\ref{eq: efe_entropy})} \label{ssec: first_term}
Recall $p(\tilde{\theta}_{\scurr, \pi^\star} |\scurr, \pi^\star)$ is the belief over parameters of the predicted observation distribution $q(\ccvec | \pi^\star)$. To address the intractability, we can extract the current estimate parameters $\expec{\tilde{\theta}_{\scurr, \pi^\star}}$. This certainty-equivalence approximation makes $q(\ccvec | \pi^\star)$ a simple multivariate normal distribution
{\allowdisplaybreaks
\abovedisplayskip = 3.5pt
\abovedisplayshortskip = 3.5pt
\belowdisplayskip = 3.5pt
\belowdisplayshortskip = 1.5pt
\begin{align} \label{eq: ceq}
    q(\ccvec | \pi^\star) &= \int_{\tilde{\theta}_{\scurr, \pi^\star}} q(\ccvec, \tilde{\theta}_{\scurr, \pi^\star} | \pi^\star)d\tilde{\theta}_{\scurr, \pi^\star} \notag \\ 
    &\approx q(\ccvec | \pi^\star ; \expec{\tilde{\theta}_{\scurr, \pi^\star}}).
\end{align}
}
The first term can then be calculated as follows
{\allowdisplaybreaks
\abovedisplayskip = 3.5pt
\abovedisplayshortskip = 3.5pt
\belowdisplayskip = 3.5pt
\belowdisplayshortskip = 1.5pt
\begin{align}
    &-\mathbb{E}_{q(\ccvec | \pi^\star)} \Big[ \log p_{\pref}(\ccvec) \Big] \approx 
    \log((2\pi)^{-k/2} |\mathbf{\Sigma}_{\pref}|^{-1/2}) \notag \\
    &+ \frac{1}{2} \big( \mathbf{\mu}_{\pref}^T \mathbf{\Sigma}_{\pref}^{-1} \mathbf{\mu}_{\pref} 
    + \text{tr}(\mathbf{\Sigma}_{\pref}^{-1} \mathbf{\Sigma}_{\scurr, \pi^\star}) + \mathbf{\mu}_{\scurr, \pi^\star}^T \mathbf{\Sigma}_{\pref}^{-1} \mathbf{\mu}_{\scurr, \pi^\star} \big) \notag \\
    &- \mathbf{\mu}_{\pref}^T \mathbf{\Sigma}_{\pref}^{-1} \mathbf{\mu}_{\scurr, \pi^\star}, 
\end{align}
}where $\mathbf{\mu}_{\pref}$ and $\mathbf{\Sigma}_{\pref}$ are the mean vector and the covariance matrix of the user's preference distribution (see Appendix \ref{appendix: efe_pareto_derivation_first} for detailed derivation). 

\subsubsection{\edt{Calculating} the Second Term of (\ref{eq: efe_entropy})} \label{ssec: second_term}
The convolved distribution $p(\tilde{\theta}_{\scurr, \pi^\star} | \scurr, \pi^\star)$ is analytically intractable. However, we can leverage the CLT to \edt{determine a suitable proposal distribution $q(\cdot)$ that closely models $p(\cdot)$}.  
Since each $p(\tilde{\theta}_{s,a})$ is NIW-distributed, 
both the mean and variance are well-defined when $\kappa > 0$ and $\nu > N + 1$. 
Therefore, under the CLT, $p(\tilde{\theta}_{\scurr, \pi^\star} | \scurr, \pi^\star)$ tends towards a MVN distribution as the plan length $|\pi^\star|$ becomes large \cite{dunn2022exploring}.

Consider the following vectorization of the parameters $\theta_{s,a}$
\begin{equation} \label{eq: vectorization}
    \text{vec}(\theta_{s,a}) = (\mu_1, \ldots, \mu_N, \sigma^2_{1,1}, \ldots, \sigma^2_{N,N}),
\end{equation}
where $\mu_1, \ldots, \mu_N$ represent the components of $\mu_{s,a}$, and $\sigma^2_{1,1}, \ldots, \sigma^2_{N,N}$ are the unique upper triangular elements of $\mathbf{\Sigma}_{s,a}$. In total, the vectorization has dimension ${N(N+3)}/{2}$. 
\edt{Due to the linearity of expectation and independence between state-action pairs,
\begin{align} 
    \expec{\tilde{\theta}_{s_K, \pi}} &= \sum_{k=0}^m \expec{\tilde{\theta}_{s_k, a_k}}, \label{eq: expec_params_1} \\
    \variance{\tilde{\theta}_{s_K, \pi}} &= \sum_{k=0}^m \variance{\tilde{\theta}_{s_k, a_k}}. \label{eq: expec_params_2}
\end{align}
}\edt{Therefore,} the distribution $p(\tilde{\theta}_{\scurr, \pi^\star} | \scurr, \pi^\star)$ can be approximately represented by multivariate normal \edt{proposal} distribution parameterized by the vectorized cumulative mean and variance
\edt{\begin{align} \label{eq: moment_match}
    q(\tilde{\theta}_{\scurr,\pi^\star}\!| \scurr,\!\pi^\star\!)\!&\equiv\! \mathcal{N}\!\big(\!\sum_k\expec{\text{vec}(\tilde{\theta}_{s_k,a_k})},\! \sum_k\!\variance{\text{vec}(\tilde{\theta}_{s_k,a_k})}\big).
\end{align}}Since $q(\tilde{\theta}_{\scurr, \pi^\star} | \scurr, \pi^\star)$ is MVN, the entropy of the proposed distribution (second term) can be represented analytically rewritten as follows.
{\allowdisplaybreaks
\abovedisplayskip = 3.5pt
\abovedisplayshortskip = 3.5pt
\belowdisplayskip = 3.5pt
\belowdisplayshortskip = 1.5pt
\begin{align}
    \entropy{q(\tilde{\theta}_{\scurr, \pi} |\scurr, \pi^\star)} \edt{=} & 
    \frac{1}{2}\log \big( \text{det}\big(\sum_k\variance{\text{vec}(\tilde{\theta}_{s_k,a_k})} \big) \big)\notag \\ &+ \frac{N(N+3)}{4}(1 + \log 2\pi).
\end{align}
}

\subsubsection{Approximating the Third Term of (\ref{eq: efe_entropy})} \label{ssec: third_term}
Using both the certainty equivalence approximation in (\ref{eq: ceq}) and the CLT approximation in (\ref{eq: moment_match}), the remaining third term can be approximated using Monte Carlo sampling of the expectation. 
To sample the expectation, a cumulative cost vector must be sampled from $q(\ccvec |  \pi^\star)$. Using (\ref{eq: ceq}), this sample (denoted $\hat{\cvec}$) can be drawn from the MVN $\hat{\ccvec} \sim q(\ccvec | \pi^\star ; \expec{\tilde{\theta}_{\scurr, \pi^\star}})$. However, note that the posterior $p(\tilde{\theta}_{\scurr, \pi^\star} | \ccvec, \scurr, \pi^\star)$ is itself a convolved NIW conditioned on $\cvec$. To construct an analytical approximation of the posterior, we must instead expand the sampling procedure to each state-action distribution.
Instead of sampling $\hat{\ccvec}$ directly from $q(\ccvec | \pi^\star ; \expec{\tilde{\theta}_{\scurr, \pi^\star}})$, $\hat{\cvec}$ can instead be constructed from the observation distribution for each state action pair
\begin{align}
    \hat{\ccvec} = \sum_k \hat{\cvec}_{s_k, a_k},
\end{align}
where $\hat{\cvec}_{s_k, a_k}\!\sim\!p(\cvec | s_k, a_k; \expec{\tilde{\theta}_{s_k,a_k}})$.
For each 
sample $\hat{\cvec}_{s_k,a_k}$, the respective posterior 
$p(\tilde{\theta}_{s_k,a_k} | \hat{\cvec}_{s_k,a_k})$ are computed using the equations found in Sec. \ref{sec: update}. The distribution $p(\tilde{\theta}_{s_k,a_k} | \hat{\cvec}_{s_k,a_k})$ is the conjugate NIW posterior.
Therefore, using the same process described in Sec. \ref{ssec: second_term}, an approximation of 
$\entropy{q(\tilde{\theta}_{\scurr, \pi} |\hat{\ccvec}, \scurr, \pi^\star)}$ can be calculated. \edt{This sampling procedure is repeated $n_s$ times to compute the third term of (\ref{eq: efe_entropy}). Using the aforementioned techniques, we can optimize for $\pi^\star_{\pref}$ in \eqref{eq:argminefe}. While this sampling procedure is computationally burdensome, the selection procedure runs in $O(n_s m|\tilde{\Pi}^\star|)$, linear in the size of number of candidate Pareto optimal plans.}

\edt{Note that unlike the prior distribution in the second term of \eqref{eq: efe_entropy}, the proposal posterior $q(\tilde{\theta}_{\scurr, \pi^\star} | \ccvec,\!\scurr,\!\pi^\star)$ is not a variational proposal distribution. In fact, the approximation of true posterior $p(\tilde{\theta}_{\scurr, \pi^\star} | \ccvec,\!\scurr,\!\pi^\star)$ with $q(\tilde{\theta}_{\scurr, \pi^\star} | \ccvec,\!\scurr,\!\pi^\star)$ is an approximation \textit{inherent} to active inference \cite{smith2022}, as $p(\tilde{\theta}_{\scurr, \pi^\star} | \ccvec,\!\scurr,\!\pi^\star)$ is often not analytically tractable. This approximation intuitively relies on the assumption that the internal model of the agent is accurate enough to well predict the true posterior. 
Our empirical evaluations in Appendix \ref{appendix: mvn_sub} show that the approximation of convolved NIW with convolved MVN agrees with the CLT. That is, the approximation is fairly accurate for the typical plan length seen in the experiments in Sec. \ref{sec: experiments}, and increases as the plan length increases (Fig. \ref{fig:plan_length_qq}) and more data is collected (Fig. \ref{fig:data_qq}). Additionally, refer to Appendix \ref{appendix: mc_sampling} for an error analysis of the Monte Carlo sampling procedure described in Sec. \ref{ssec: third_term} with respect to the number of samples $n_s$, plan length, and collected data.
}

\subsection{Execution and Parameter Update} \label{sec: update}

After obtaining the preferred optimal trade-off plan $\pi^\star_{\pref}$, it is executed on the robotic platform (or in simulation). During the execution, measurements for the cost of each state-action $\hat{\cvec}$ are collected. Recall, the 
cost parameters 
are distributed as $p(\tilde{\theta}_{s,a} | D_{s,a}) = NIW_{s,a}(\lambda, \kappa, \Lambda, \nu)$. Given a set of measurements $D_{s,a} = \{\hat{\cvec}_0, \ldots \hat{\cvec}_l \}$, the conjugate posterior is a NIW parameterized as follows. 
%
\begin{align}
    \lambda &= \frac{\kappa_0 \lambda_0 + l \bar{\cvec}}{\kappa_0 + l},\\
    \kappa &= \kappa_0 + l,\\
    \Lambda &= \Lambda_0 + \frac{\kappa_0 l}{\kappa_0+ l}(\bar{\cvec}-\lambda_0)(\bar{\cvec}-\lambda_0)^T + \\ 
    & \qquad \sum_{j=0}^l(\hat{\cvec}_j - \bar{\cvec})(\hat{\cvec}_j - \bar{\cvec})^T, \nonumber \\
    \nu &= \nu_0 + l,
\end{align}

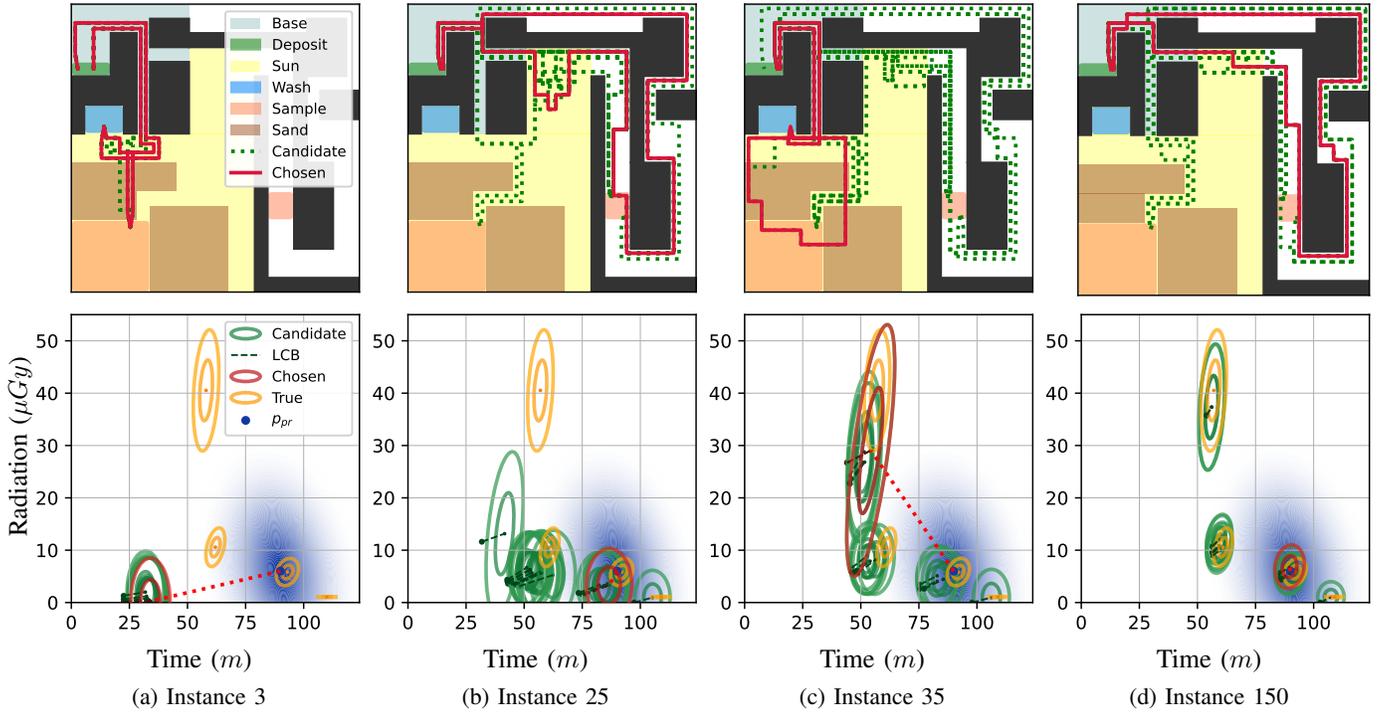
\begin{figure*}
    \centering
    
    \begin{subfigure}[t]{0.24\textwidth}
        \centering
        \includegraphics[width=1\linewidth]{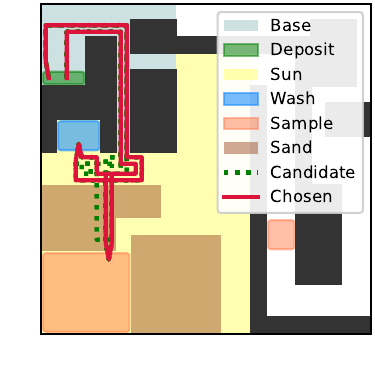}
    \end{subfigure}
    \begin{subfigure}[t]{0.24\textwidth}
        \centering
        \includegraphics[width=1\linewidth]{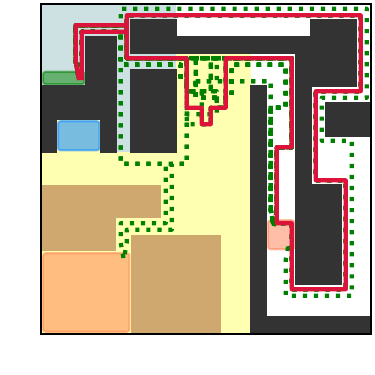}
    \end{subfigure}
    \begin{subfigure}[t]{0.24\textwidth}
        \centering
        \includegraphics[width=1\linewidth]{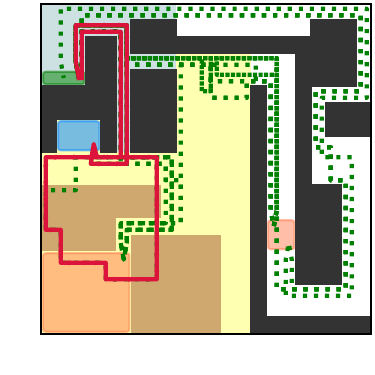}
    \end{subfigure}
    \begin{subfigure}[t]{0.24\textwidth}
        \centering
        \includegraphics[width=1\linewidth]{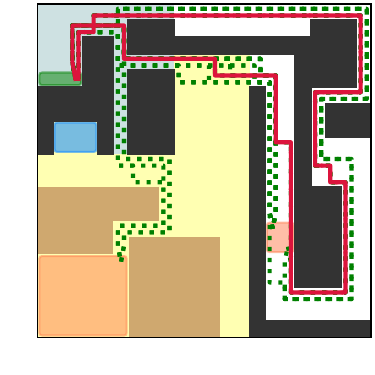}
    \end{subfigure}

    \vspace{2mm}
    \vspace{-\baselineskip}
    
    \begin{subfigure}[t]{0.24\textwidth}
        \includegraphics[width=1\linewidth]{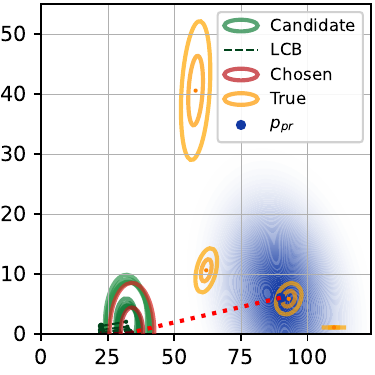}
         \begin{tikzpicture}[remember picture, overlay]
             \node[inner sep=0pt, anchor=center] (graphic) at (0,0) {};
             \node[below] at (2.18, 0.4) {Time ($m$)};
             \node[left, rotate=90] at (-.2, 4.3) {Radiation ($\mu Gy$)};
         \end{tikzpicture}
         \caption{Instance 3}
        \label{fig:instance_3}
    \end{subfigure}
    \begin{subfigure}[t]{0.24\textwidth}
        \centering
        \includegraphics[width=1\linewidth]{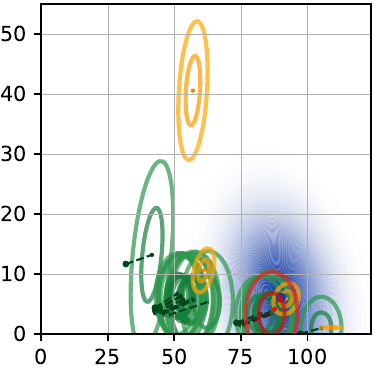}
         \begin{tikzpicture}[remember picture, overlay]
             \node[inner sep=0pt, anchor=center] (graphic) at (0,0) {};
             \node[below] at (0, 0.4) {Time ($m$)};
         \end{tikzpicture}
        \caption{Instance 25}
        \label{fig:instance_25}
    \end{subfigure}
    \begin{subfigure}[t]{0.24\textwidth}
        \centering
        \includegraphics[width=1\linewidth]{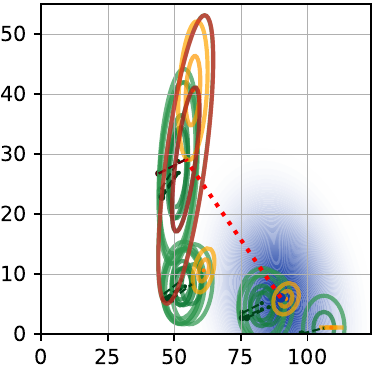}
         \begin{tikzpicture}[remember picture, overlay]
             \node[inner sep=0pt, anchor=center] (graphic) at (0,0) {};
             \node[below] at (0, 0.4) {Time ($m$)};
             \node[below] at (0, 0.4) {};
         \end{tikzpicture}
        \caption{Instance 35}
        \label{fig:instance_35}
    \end{subfigure}
    \begin{subfigure}[t]{0.24\textwidth}
        \centering
        \includegraphics[width=1\linewidth]{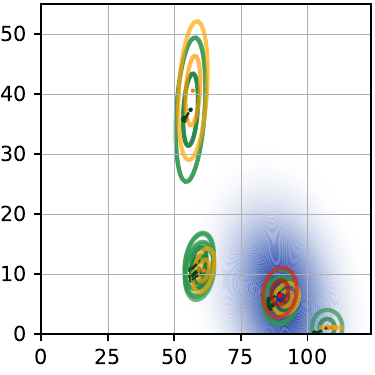}
         \begin{tikzpicture}[remember picture, overlay]
             \node[inner sep=0pt, anchor=center] (graphic) at (0,0) {};
             \node[below] at (0, 0.4) {Time ($m$)};
             \node[below] at (0, 0.4) {};
         \end{tikzpicture}
        \caption{Instance 150}
        \label{fig:instance_150}
    \end{subfigure}
    
    \caption{Simulated rover sample collection: A rover is tasked with repetitively collecting a $sample$ and delivering it to $deposit$. Moving takes time and collects radiation (in the sun). The user prefers that the sample is collected in roughly 90 minutes with about 6 $\mu Gy$ of radiation, represented with $p_\pref$. 
    Top row figures show the computed plans, and the bottom row figures show the estimated Pareto points.
    The true optimal trade-off plans are shown for comparison in four instance (episode) snapshots. \textit{Video:} \url{https://youtu.be/tCRJwqeT-f4}. 
    }
    \label{fig:rover}
\end{figure*}

By iteratively performing each described phase, the agent can safely complete the task, guided by the user's preference.
Through the iterative combination of exploratory planning, surprised-based selection, execution, and Bayesian update of cost distributions, the proposed MORL framework can achieve all three goals described in Prob. \ref{problem}. Now, we evaluate the empirical efficacy through experiments.

\section{Experiments} \label{sec: experiments}
To evaluate the effectiveness of our MORL framework for autonomous robotic decision making in unknown environments, we performed three case studies: (i) an illustrative simulation study, (ii) two numerical benchmarking experiments for comparison against the state-of-the-art, and (iii) a hardware experiment to demonstrate the real-world applicability of the method\footnote{To see a video of the simulation and hardware experiments, visit \url{https://youtu.be/tCRJwqeT-f4}}. 

\subsection{Simulated Mars Surface Exploration Study}
\subsubsection{Motivation} \label{sec: rover_setup}
Suppose a Mars rover is tasked to collect scientifically interesting minerals from target sample sites (orange region in Fig. \ref{fig:rover}), designated based on past mission data~\cite{farley2020mars}. Due to the limited number of sample tubes, it is required to go back and forth between the deposit location (green region) and the target site. 
Since the closer sampling region is in the sun, the rover must trade-off between low-time, high-radiation for efficient science data collection, or high-time, low-radiation to avoid long-term damage.
Therefore, to protect itself while gaining sufficient amount of science, it must strike a balance between the two. Moreover, although the rover knows the qualitative information about the environment (e.g., where sand, base, washing station, etc. are located), it is still necessary to estimate online the time and radiation cost of collecting a sample. Informed by past mission data and scientific expertise, the user can tune an appropriate $p_{\pref}$.

\subsubsection{Simulation Setup} For the sake of simplicity, the rover can travel in cardinal directions, and may not move through obstacles (black). The task specification is given as 
\begin{multline*}
    \phi_{\text{MR}} = F(sample \, \land \, F(deposit)) \, \land \\ G(sand \rightarrow (\neg base \, U wash));
\end{multline*}
in English, ``\textit{collect a sample, then deposit it, and, if sand is visited then wash before returning to base}.'' Each transition takes time (objective 1) measured in minutes, and transitions in the $sun$ accumulate harmful radiation (objective 2) measured in micrograys. The true mean costs for transitions in each region are given as follows: 
\begin{align*}
     &\mu_{sample(left)} = (6,16) && \text{(in the sun)}, \\
     &\mu_{sample(right)} = (6,0) && \text{(out of the sun)}, \\
     & \mu_{sand} = (3,7) && \text{(slows the rover in the sun)}, \\
     &\mu_{wash} = (11,31) && \text{(long duration of radiation exposure)},
\end{align*}
$\mu_{sun} = (1,1)$, and all other transitions have $\mu = (1,0)$.
Covariance values are omitted for brevity.
The user informs the prior transition cost for all transitions to be $\lambda_0 = (0.5, 0)$. For the longevity of the rover, the user aims to endure a small amount of radiation and prefers that the mission is completed in roughly one and a half hours. To express this, the chosen preference distribution is 
\begin{gather*}
    \N(\mean_\pref, \cov_\pref), \quad \mean_{\pref} = (90, 6) \\
    \sigma_{1,1} = 140, \;\;\; \sigma_{1,2}=\sigma_{2,1} = -2, \;\;\; \sigma_{2,2} = 70, 
\end{gather*}
where $\sigma_{i,j}$ is the $i$-$j$ element of $\cov_\pref$.
Soon after deployment (instance 3), the planner only finds candidate plans that go through sand to collect the sample, and then must wash off, seen in Fig.~\ref{fig:instance_3}. The robot quickly learns to avoid sand as washing takes very long and collects lots of radiation. The robot gradually discovers more optimal behavior seen by instance 25 in Fig. \ref{fig:instance_25}. In instance 35, seen in Fig. \ref{fig:instance_35}, the rover has roughly learned the true Pareto front, and elects to explore a less preferred, information rich trade-off. Finally, after 150 instances, seen in Fig. \ref{fig:instance_150}, the robot has obtained a fairly accurate estimate of the true Pareto front, and has converged to often selecting the most preferred trade-off.

\subsection{Benchmarks}

We evaluate the performance of our framework against the state-of-the-art in two benchmark problems.
The goal is to assess
in each instance: (i) how close the selected plan $\pi^\star_{\pref}$ is to being Pareto optimal, and (ii) how well $\Pi^\star$ is represented by $\tilde{\Pi}^\star$. Metric \emph{cumulative Pareto-regret} directly evaluates (i). To properly evaluate (ii), 
we define \emph{Pareto-bias} metric

\begin{figure*}[htbp]
    \centering
    
    \begin{subfigure}[t]{0.32\textwidth} 
        \centering
        \includegraphics[trim={3mm 2mm 4.2mm 2mm},clip,width=1\linewidth]{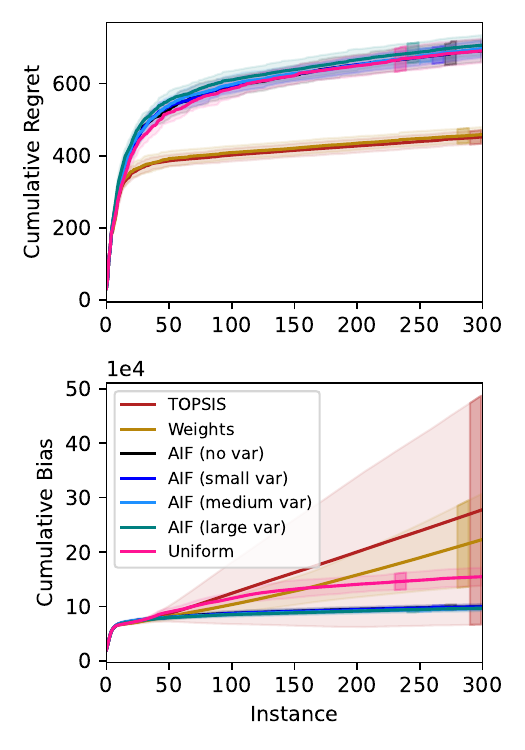}
        \caption{10 true Pareto points}\label{fig: fixed_few}
    \end{subfigure}
    \begin{subfigure}[t]{0.32\textwidth} 
        \centering
        \includegraphics[trim={3mm 2mm 4.2mm 2mm},clip,width=1\linewidth]{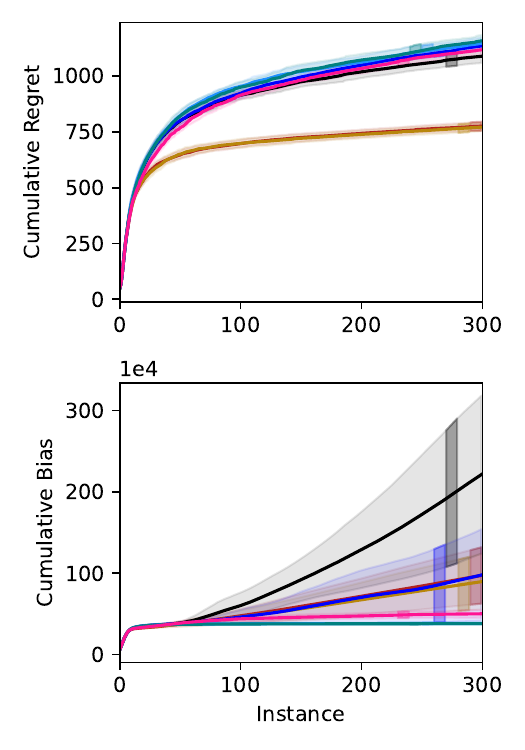}
        \caption{32 true Pareto points}\label{fig: fixed_some}
    \end{subfigure}
    \begin{subfigure}[t]{0.32\textwidth} 
        \centering
        \includegraphics[trim={3mm 2mm 4.2mm 2mm},clip,width=1\linewidth]{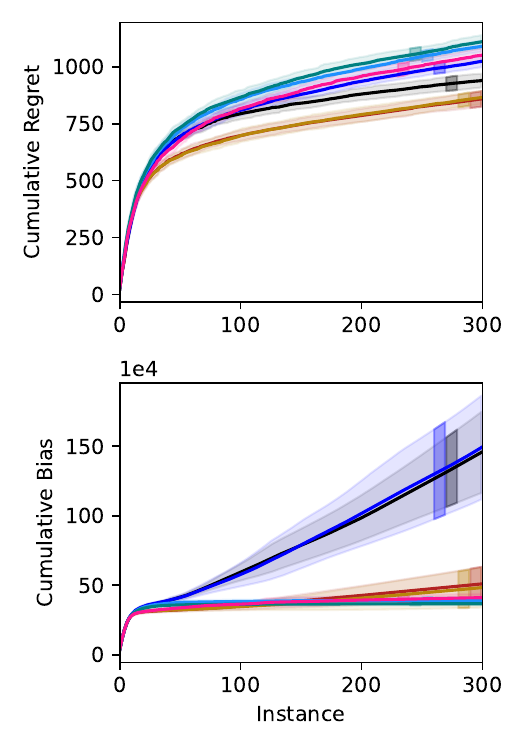}
        \caption{72 true Pareto points}\label{fig: fixed_many}
    \end{subfigure}
    
    \caption{Specific environment benchmark: Within our framework, we benchmark our active inference (AIF) selection method against other state-of-the-art methods with respect to cumulative Pareto-regret (top) and Pareto-bias (bottom) for three fixed environments with varying numbers of true Pareto points. The mean is represented by the solid line, with 1-$\sigma$ variance shown by the shaded region. \edt{The dark shaded bar aids in distinguishing overlapping variance bands.}}
    \label{fig:fixed_benchmarks}
    \vspace{-4mm}
\end{figure*}

for a true Pareto front $\Omega$ and an estimate Pareto front $\tilde{\Omega}$ as
\begin{equation} \label{eq: pareto_bias}
        B\!=\!\frac{1}{|\Omega|}\!\sum_{\ccvec_i \in \Omega} \!\min_{\ccvec_j \in \tilde{\Omega}}(d(\ccvec_i, \!\ccvec_j))\!+\!\frac{1}{|\tilde{\Omega}|}\!\sum_{\ccvec_i \in \tilde{\Omega}} \min_{\ccvec_j \in \Omega}(d(\ccvec_j, \!\ccvec_i)), 
\end{equation}
where 
\begin{align*}
    d(\ccvec_i,\!\ccvec_j)\!=\!\mathcal{W}_2\!\Big(\!p(\ccvec_i|\scurr,\!\pi_i;\! \theta_{\scurr, \pi_i}\!),\! q(\ccvec_j | \scurr,\!\pi_j;\! \expec{\tilde{\theta}_{\scurr, \pi_j}})\!\Big)
\end{align*}
is the Wasserstein-2 ($\mathcal{W}_2$) distance.
The first term in \eqref{eq: pareto_bias} quantifies the estimate Pareto front's coverage of the true Pareto front, and the second term penalizes outlier/excess trade-offs in the estimate Pareto front. If each true Pareto point is exactly similar to an estimate Pareto point, and visa versa, then $B\!=\!0$, otherwise $B\!>\!0$.

We benchmarked our active inference (AIF) selection method against other state-of-the-art selection methods: uniform random selection (fair selection), TOPSIS \cite{mirzanejad2022} (objective preference), and linear scalarization, i.e. ``weights'' (subjective preference). Refer to Sec. \ref{sec: background} for further description of the compared methods.

\subsubsection{Fixed Environment Case Study}
Three grid world environments were designed to challenge the learning of different size Pareto fronts. All experiments were run for 300 instances, with a LCB confidence parameter $\alpha=0.1$, \edt{and $n_s=300$ Monte Carlo samples, selected via cross validation. See Appendix \ref{appendix: mc_sampling} for more details on the accuracy with respect to the choice of $n_s$.} Fig. \ref{fig:fixed_benchmarks} shows benchmarks against each environment. Generally, methods that accumulate less bias require learning more of the transition cost model via exploration, and hence accumulate more regret. When learning the small Pareto front, seen in Fig. \ref{fig: fixed_few}, AIF performed better than both uniform and the biased methods (weights and TOPSIS) in terms of bias, with comparable regret to uniform. This is due to the advantageous information-gain-based exploration. When the number of Pareto points on the Pareto front increases, seen in Fig. \ref{fig: fixed_some}, using {\it small} and {\it no}
\footnote{If the user only wants to specify a desired point in objective-space, they can make the preference distribution covariance very small (\textit{no} variance)} 
variance selects for only a portion of the true Pareto front, while the {\it medium} and {\it large} variance still envelope the entire Pareto front. 
With the smaller variance AIF methods, once a good estimate candidate is found, the first term of (\ref{eq: efe_entropy}) dominates the selection, and thus abandons the less promising options that are beyond the variance. With an even larger Pareto front seen in Fig. \ref{fig: fixed_many}, this effect is exacerbated. In this case, the {\it no} variance method sees a decrease in regret due to the heavy myopic bias after finding a candidate solution that agrees with the preference. 
Thus, from this experiment, we can observe that AIF methods start becoming very selective once the preferred region (which can be varied by the size of variance) on the Pareto front is discovered.

\subsubsection{Random Environment Case Study}
The behavior of our AIF selection method can vary widely across different scenarios depending on the diversity of the Pareto front relative to $p_\pref$. Therefore, we illustrate the general behavior of our approach using randomized grid worlds. Each trial, a 20$\times$20 grid world robot model is generated with randomized proposition locations and true cost distribution parameters. The data was analyzed across 750 randomized trials.

\begin{figure}
    \centering
    \includegraphics[trim={3mm 2mm 4.2mm 2mm},clip,width=0.38\textwidth]{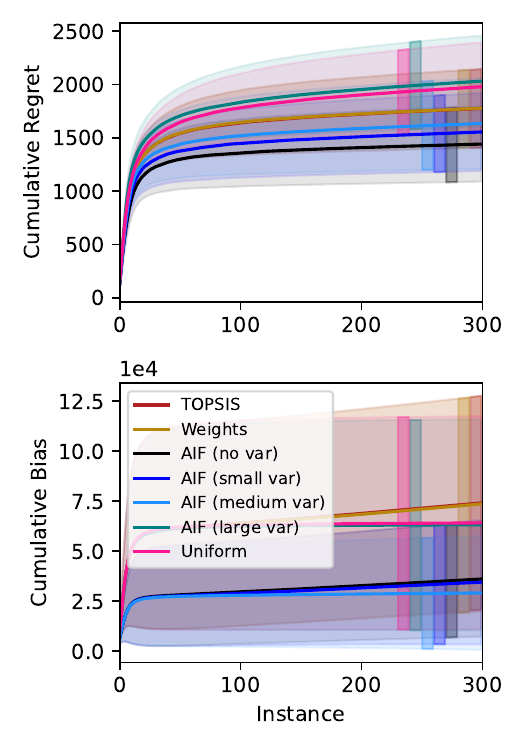}
    \caption{Randomized environment benchmark: We benchmark our method against other selection methods for randomly generated scenarios of varying complexity.}
    \label{fig:rand_benchmarks}
    \vspace{-4mm}
\end{figure}

Fig. \ref{fig:rand_benchmarks} demonstrates the general trade-off between bias and regret. Note that the standard deviation in performance among trials is very large, since each randomly generated environment is diverse. When the variance of $p_{\pref}$ is {\it large}, active inference tends to perform similarly to random selection (Uniform) in terms of both regret and bias. With a moderate variance, active inference tends to have better regret and bias performance than other methods due to the intelligent information theoretic exploration when the estimate plans $\tilde{\Pi}^\star$ are uninformed. \edt{Uniform, AIF (large var), and AIF (medium var) stabilize in cumulative bias, meaning they effectively learn the entire Pareto front. The remaining methods converge to a non-constant slope cumulative bias, indicating that portions of the Pareto front have not been properly learned.} From this experiment, we can observe that AIF methods outperform other state-of-the-art methods in terms of both cumulative regrets and biases as long as the size of variance is not too large.  

\edt{Akin to other expected free energy approaches, the enhanced reasoning power comes at a computational cost compared to other methods \cite{kaplan2018, smith2022}. 
Uniform selection has complexity $O(1)$, weights has $O(|\tilde{\Pi}^\star|)$, and TOPSIS has $O(|\tilde{\Pi}^\star|^2)$ \cite{hamdani2016complexity}.
We benchmarked the aforementioned random environment case studies with respect to both \textit{planning} time and \textit{selection} time per instance on a computer with AMD Ryzen 7 3800X 3.9 GHz 8-Core Processor. The average planning time is across 400 experiments with 100 instances each is $483 \pm 443$ms (1-$\sigma$) with a median planning time of $369$ms. The average AIF selection time (across all four preference distribution variances) is $610 \pm 1051$ms (1-$\sigma$) with a median selection time of $179$ms. The compared selection procedures completed in less than $1$ms.}

\edt{
\begin{remark}
    For larger-scale robotic models, the exact multi-objective graph search can be replaced with approximate multi-objective search, such as A$^*$pex \cite{zhang2022pex}, to admit a smaller (approximate) Pareto front in less time. Additionally, a clustering algorithm \cite{zio2011clustering, bejarano2022clustering} can further reduce the size of the Pareto front. We expect both adaptations to our framework to greatly enhance computation speed at the cost of Pareto-regret and bias performance, however we leave this to future work.
\end{remark}
}


%

\subsection{Hardware Experiment}
\begin{figure*}[t]
    \centering
    \includegraphics[trim={0 25mm 0 0},clip,width=0.9\linewidth]{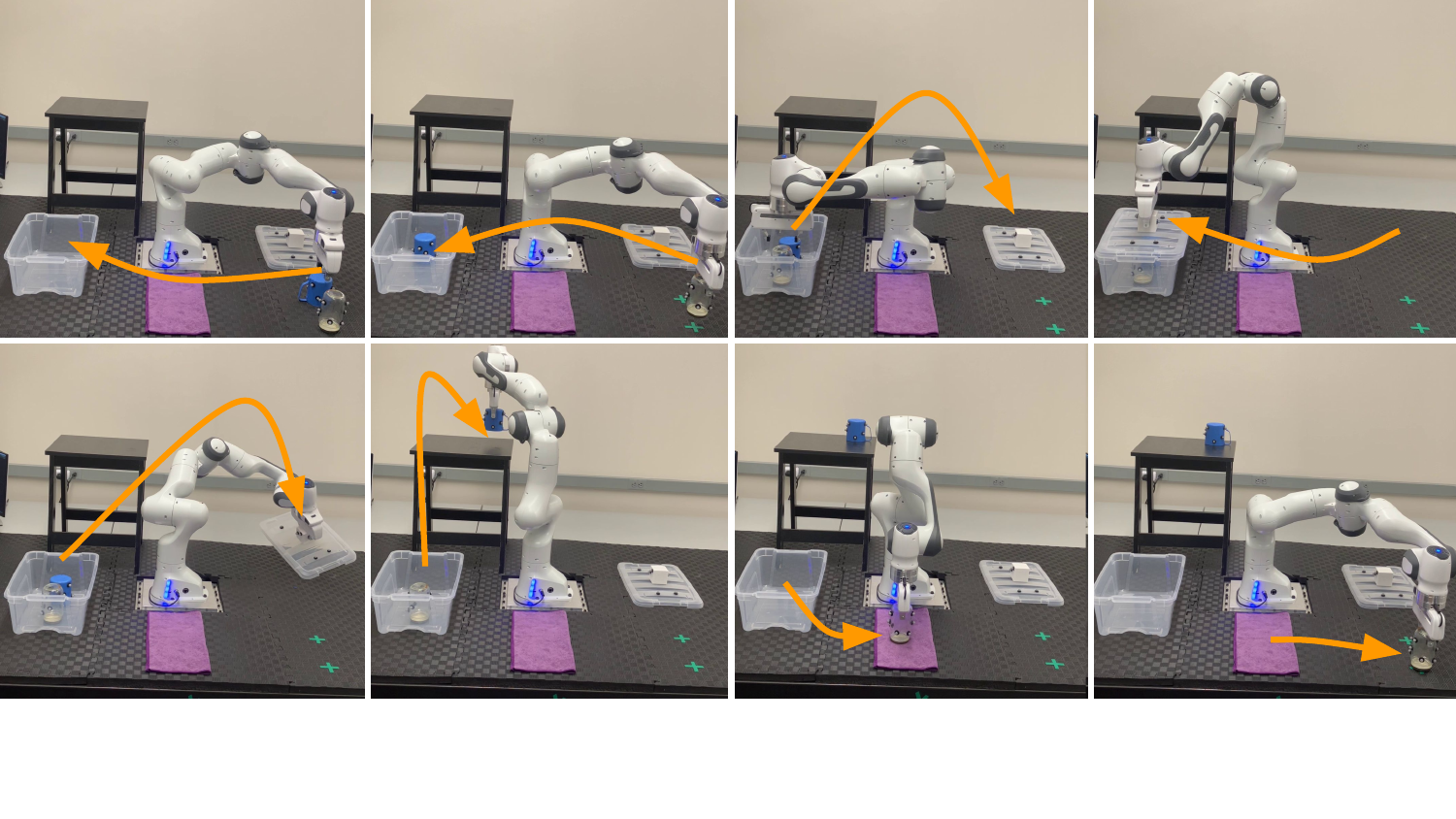}
    \caption{Hardware dishwashing experiment: A robotic manipulator must load both a durable pitcher and a fragile glass jar into a dishwasher, then must unload each item by either placing it on the drying rack, or manually drying it on the ground. The robot learns to minimize time and risk (holding the jar high above the ground). The user prefers a low-risk, higher-time trade-off. After $500$ instances, the system performs optimally (left to right, top to bottom), and decides to quickly place the pitcher on the rack, and spend time to manually dry the jar. \textit{Video:} \url{https://youtu.be/tCRJwqeT-f4}. }
    \label{fig: hw_experiment}
    \vspace{-0.2in}
\end{figure*}

To demonstrate the efficacy and diversity of application of our framework, we studied a complex dishwashing experiment using a robotic manipulator described in Fig. \ref{fig: scenario}. The robot must repetitively load and unload the dishwasher with two items, a durable pitcher, and a fragile glass jar. The LTLf specification is 
$$\phi = \phi_l \land \phi_s,$$
where
\begin{align*}
    \phi_l =& F(P_{dw} \land J_{dw} \land lid_{on} \land F(P_{rack} \lor (P_{dry} \land F P_{floor}) \land \\
    &F(J_{rack} \lor (J_{dry} \land F J_{floor}))),  \\
    \phi_s =& G(\neg lid_{aside} \rightarrow F(P_{dw} \land J_{dw})),
\end{align*}
which can be interpreted as ``\textit{put the pitcher and the jar in the dishwasher, then put the lid on, then for both the pitcher and jar, place on the drying rack, or dry the item and place it on the ground}''. The performance of the robot is evaluated by the cumulative execution time (objective 1), as well as cumulative risk (objective 2) measured for a given action primitive,
\begin{align}
    risk(s,a) = \begin{cases}
        \int_t h_{J}^2 dt & \text{ if holding jar,}\\
        0 & \text{ otherwise,}
    \end{cases}
\end{align}
where $h_{J}$ is the height of the jar above the ground. A sampling based motion planner (PRM$^*$ \cite{lavalle2006planning}) is used to realize motion action primitives, inducing unknown and random time and risk costs. Note that placing the jar on the drying rack is likely to have high risk, whereas manually drying the jar on the ground will likely take more time. We trained the transition cost model in simulation for $500$ instances, using a preference distribution of $\mu_{\pref} = (350, 0.5)$ and $\sigma_{1,1}^2 = 400$, $\sigma_{2,2}^2 = 2$, $\sigma_1 \sigma_2 = 0$, which selects for a high-time, low-risk alternative. The converged behavior is shown in Fig. \ref{fig: hw_experiment}. As can be seen, the robot correctly loads the dishwasher, then unloads the pitcher on the rack, and decides to manually dry the jar, all while not performing any sub-optimal/unnecessary actions. The video of the execution is included in the supplementary material and visually represents the applicability of our MORL framework to a variety of robotic models with complex tasks.


\section{Related Work} \label{sec: background}


\subsection{Multi-Objective Reinforcement Learning} \label{sec: multi-objective_optimization}

The goal of MORL is to maximize a multi-objective value or utility function \cite{hayes2022}. 
MORL extends single-objective RL methods by selecting among Pareto-optimal candidate actions.
One method of selection is casting valuations of candidate actions to a single-objective by scalarizing, in turn requiring a sound and interpretable utility function. 

Scalarization of utility functions can be classified as \textit{subjective}, scalarizing based on an exogenous user's preferred trade-off \cite{abels2019}, or \textit{non-subjective}, scalarizing without external preferences \cite{wang2012, mirzanejad2022}. 
Linear scalarization is a standard subjective method that combines objectives objectives through a relative-importance weighted sum.
On the contrary, non-subjective methods such as TOPSIS \cite{mirzanejad2022} and hyper-volume indicators \cite{wang2012} rely on Euclidian distance metrics in objective space. However, in many robotic applications where objectives have different units of measurement (e.g. time in seconds and energy in joules), these methods use the addition of quantities with different units, which is generally mathematically ad-hoc and lacks interpretation.
Linear scalarization can potentially combat this issue by interpreting weights as reciprocal measures (e.g. $1/$second), at the cost of requiring a user to have a rigorous understanding of how to compare the importance of each objective and properly assign weights. Furthermore, re-interpreting units of measurement (e.g. seconds instead of milliseconds) alters Euclidian distance, which can dramatically alter the selected solution, when no fundamental aspects about the problem have changed.
On the contrary, our method avoids this issue by only describing the scalarization through operations on probability density functions.

Many MORL methods do not distinguish between optimizing multi-objective value function and selecting the best action from the Pareto-optimal candidate actions \cite{abels2019, hayes2022}. Instead, these methods scalarize using a monotonic utility 
function, which guarantees that an optimal (preferred) action with respect to the single-objective utility function must also be Pareto-optimal in objective space. The use of such a utility function effectively optimizes a single objective, which casts a complete order on the reward space and eliminates the need for explicit computation of the Pareto front beforehand. However, when a non-monotonic utility function is used, or the agent is concerned with learning the set of all optimal trade-off solutions \cite{drugan2013}, the Pareto front must be explicitly computed before \textit{selection}, as seen in many MOMAB approaches.
Our framework computes the Pareto front using a multi-objective task-planner.

MOMAB problems are concerned with minimizing \textit{Pareto-Regret} \cite{drugan2013}, which uses $\varepsilon$\textit{-dominance} to measure how ``far'' the chosen action is from being Pareto-optimal with respect to the true unknown reward distributions. 
Besides Pareto-regret, the MOMAB performance can also be evaluated by \textit{fairness}, which quantifies how evenly Pareto-optimal actions were selected. Fairness is used to compare MOMAB selection by diversity in selection and learning of all actions on a possibly non-convex Pareto-front. To learn the entire Pareto front, work \cite{drugan2013} extends the single-objective Upper Confidence Bound (UCB) algorithm \cite{auer2002} to multiple objectives, then rely on random selection among the Pareto front. While random selection effectively learns the entire Pareto front, it cannot account for or converge to a preferred trade-off.
To address this shortcoming, our formulation of the high-level active inference decision making agent naturally balances this exploration vs. exploitation trade-off.

\subsection{Active Inference and Expected Free Energy} \label{sec: active_inference}
Active inference is an uncertainty-based sequential decision making scheme that applies the free energy principle \edt{(FEP) \cite{friston2010}} to the behavioral norms of physical agents \cite{friston2015,kaplan2018,friston2022}. 
\edt{According to the FEP, agents intend to minimize the surprise (in our problem it is denoted as $-\log p(\ccvec|\pi)$) to maintain their homeostases. However, calculating the surprise directly requires to sum/integrate over all possible states/parameters in their internal generative models. When these quantities are continuous, depending on the prior and/or the likelihood function, the integral does not have a closed-form solution. Thus, some approximation methods are required to calculate the surprise.}  

\edt{One of the approaches is to employ variational Bayesian inference in statistical machine learning \cite{bishop2006}. The core idea of this methodology is to introduce a proposal/surrogate distribution, which is {\it easily} modeled, and optimize the functional of this distribution to make it resemble a true intractable distribution. More specifically, in our problem, the following free-energy (FE, i.e. negative evidence lower bound) is minimized such that the Kullback-Leibler Divergence (KLD) between the proposal distribution $q(s,\tilde{\theta}|\pi)$ and the true distribution $p(s,\tilde{\theta}|\ccvec, \pi)$ approaches zero (Note that here we used the simplified notations seen in Appendices \ref{appendix: efe_pareto_derivation} and \ref{appendix: efe_pareto_derivation_first}).
\begin{align}
    FE &= \int_{s,\tilde{\theta}} q(s,\tilde{\theta}|\pi) \log \frac{q(s,\tilde{\theta}|\pi)}{p(s,\tilde{\theta}, \ccvec|\pi)} d\tilde{\theta} ds 
    \nonumber \\
    &= \int_{s,\tilde{\theta}} q(s,\tilde{\theta}|\pi) \log \frac{q(s,\tilde{\theta}|\pi)}{p(s, \tilde{\theta}|\ccvec, \pi)} d\tilde{\theta} ds - \log p(\ccvec|\pi) \nonumber \\
    &= D_{KL}\Big( q(s,\tilde{\theta}|\pi) || p(s,\theta|\ccvec, \pi) \Big) - \log p(\ccvec|\pi) \nonumber \\
    &\geq \mbox{Surprise}(\pi).
\end{align}
As can be seen from this equation, since KLD is greater or equal to zero, FE is the upper bound of the surprise, i.e. minimizing FE leads to minimize the surprise.}

\edt{Computation of FE is performed given an observed outcome observation $\ccvec$, which is not known until the plan is executed. Hence, in active inference, the expected free energy (EFE) is instead minimized, and similar to FE, if the proposal and the true hidden posterior distributions match, the EFE term reduces to the following, 
\begin{align}
    EFE \rightarrow -\mathbb{E}_{q(\ccvec|\pi)}[\log p_{\pref}(\ccvec)], 
\end{align}
which is ``expected" surprise. In summary, FE allows to make the surprise minimization tractable for continuous state/parameter systems, and EFE allows for the minimization FE without knowing the received observation {\it a priori}.}

\edt{Additionally, as explained in \eqref{eq: efe},} the EFE comprises of an utility term governed by a prior 
agent's desired 
observation distribution, and 
an information gain term that evaluates how much a candidate action would reduce the uncertainty of hidden states/parameters. Hence, by minimizing EFE, 
the agents can naturally balance between two modes of behavior that are essential for autonomous decision making under uncertainty; {\it exploitation} (i.e. focusing on the current best action) and {\it exploration} (i.e. trying less executed actions). Furthermore, the degree of preference for either mode can be easily adjusted by varying the prior preference distribution. For instance, using a prior preference with a large variance will cause the agents to provide more priority to exploration, and vice versa. These characteristics are particularly useful when transparent agent behavior is desired \cite{wakayama2023observationaugmented}. Using active inference for \textit{selection} marries the benefits of \textit{preferred} versus \textit{fair} selection described in Sec. \ref{sec: multi-objective_optimization} through localized information-theoretic exploration around the prior preference distribution. Additionally, specifying a preference distribution over desired trade-offs may be particularly intuitive, explainable, and even statistically informed with past data.
Throughout this work, we study the benefits of using EFE in active inference for Pareto point selection in MORL with two statistical techniques when an intractable convolved NIW is used to represent the hidden parameters.  

\section{Conclusion and Future Work} \label{sec: conclusion_future_work}
In this work we study a systematic approach to learning Pareto-optimal behavior on an unknown multi-objective stochastic cost model. In particular we examine an active inference-based approach to multi-objective decision making and we examine the resulting behavior through robotic simulation and hardware experiments, and numerical benchmarks that compare against other state-of-the-art selection methods. Our approach marries the benefits of a user-provided preferred optimal trade-off with exploration required to learn portions of the entire Pareto front. We introduce a new Pareto-bias metric that, coupled with traditional Pareto-regret, elucidates the trade-off between accurately learning the diverse Pareto front versus quickly learning a single optimal trade-off. Notably, the balance between the aforementioned behaviors can be controlled via the user's preference.

This work gives rise to a few future directions of research such as extension to stochastic transition models. 
\edt{Additionally, integrating cost-bounds for certain objectives by embedding a pruning procedure \cite{amorese2023} into the multi-objective graph search algorithm is an interesting extension. Since the costs are not known \textit{a priori}, we expect the Pareto cost \textit{upper} confidence bound to be an effective choice of pruning constraint. We are also interested in generalizing the cost distribution to a Gaussian mixture to account for multi-modality.}
Other future directions may include convergence analysis and finite time regret analysis, for example, adapting the Upper Credible Limit algorithm \cite{reverdy2014modeling} for Gaussian multi-armed bandits to the multi-objective setting. 



\bibliographystyle{unsrt}
\bibliography{root.bib}

\appendices

\section{Derivation of EFE for the Pareto Point Selection} \label{appendix: efe_pareto_derivation}
As mentioned in Sec. \ref{sec: selection}, 
marginalizing the joint distribution 
is analytically intractable. Therefore, its bound, i.e. free energy, is minimized instead. Yet, the outcomes $\ccvec$ cannot be observed until a plan $\pi^\star$ is actually executed, so the agent ends up minimizing EFE. Hereafter, for the sake of simplicity, we denote $s_{K}$ as $s$, $\pi^\star$ as $\pi$, and $\tilde{\theta}_{s,\pi}$ as $\tilde{\theta}$. 
{\allowdisplaybreaks
\abovedisplayskip = 3.5pt
\abovedisplayshortskip = 3.5pt
\belowdisplayskip = 3.5pt
\belowdisplayshortskip = 1.5pt
\begin{align} \label{eq: efe_appendix_general}
    &\mbox{EFE}(\pi) \nonumber \\
    &\!=\!\int_{s,\tilde{\theta}}\!q(s,\!\tilde{\theta}|\pi)\!\int_{\ccvec}\! q(\ccvec|s,\!\tilde{\theta},\!\pi)\!\log\!\frac{q(s,\!\tilde{\theta}|\pi)}{p(s,\!\tilde{\theta}|\ccvec,\!\pi)p_{\pref}(\ccvec)} d\ccvec d\tilde{\theta} ds,
\end{align}
}where $p_{\pref}(\ccvec)$ is the prior preference for outcomes and $q(s,\tilde{\theta}|\pi)$ is the proposal distribution for $s$ and $\tilde{\theta}$ given a plan $\pi$. Eq. (\ref{eq: efe_appendix_general}) can be decomposed into three parts as follows. 
{\allowdisplaybreaks
\abovedisplayskip = 3.5pt
\abovedisplayshortskip = 3.5pt
\belowdisplayskip = 3.5pt
\belowdisplayshortskip = 1.5pt
\begin{align}
    (\mbox{1\textsuperscript{st}}) &\!=\!-\int_{s,\tilde{\theta},\ccvec} q(s,\tilde{\theta}|\pi) q(\ccvec|s,\tilde{\theta},\pi) \log p_{\pref}(\ccvec) d\ccvec d\tilde{\theta} ds, \nonumber \\
    &\!=\!-\int_\ccvec \log p_{\pref}(\ccvec) \Big\{ \int_{s,\tilde{\theta}} q(s,\tilde{\theta},\ccvec|\pi) d\tilde{\theta} ds \Big\} d\ccvec, \nonumber \\
    &\!=\!- \mathbb{E}_{q(\ccvec|\pi)}\Big[ \log p_{\pref}(\ccvec) \Big].
\end{align}
}
{\allowdisplaybreaks
\abovedisplayskip = 3.5pt
\abovedisplayshortskip = 3.5pt
\belowdisplayskip = 3.5pt
\belowdisplayshortskip = 1.5pt
\begin{align}
    (\mbox{2\textsuperscript{nd}}) &\!=\!\int_{s,\tilde{\theta},\ccvec} q(s,\tilde{\theta}|\pi) q(\ccvec|s,\tilde{\theta},\pi) \log \frac{q(s|\pi)}{p(s|\ccvec,\pi)} d\ccvec d\tilde{\theta} ds, \nonumber \\
    &\!=\!\int_{s,\ccvec} \Big\{\int_{\tilde{\theta}} q(s,\tilde{\theta},\ccvec|\pi) d\tilde{\theta} \Big\} \log \frac{q(s|\pi)}{p(s|\ccvec,\pi)} d\ccvec ds, \nonumber \\
    &\!=\!- \int_{\ccvec} q(\ccvec|\pi) \int_s q(s|\ccvec,\pi) \log \frac{q(s|\ccvec, \pi)}{q(s|\pi)} ds d\ccvec  , \nonumber \\
    &\!=\!- \mathbb{E}_{q(\ccvec|\pi)} \Big[ D_{KL} \Big( q(s|\ccvec, \pi) || q(s|\pi) \Big) \Big].
\end{align}
}
{\allowdisplaybreaks
\abovedisplayskip = 3.5pt
\abovedisplayshortskip = 3.5pt
\belowdisplayskip = 3.5pt
\belowdisplayshortskip = 1.5pt
\begin{align}
    &(\mbox{3\textsuperscript{rd}})\!=\!\int_{s,\tilde{\theta},\ccvec} q(s,\tilde{\theta}|\pi) q(\ccvec|s,\tilde{\theta},\pi) \log \frac{q(\tilde{\theta}|\edt{s}, \pi)}{p(\tilde{\theta}|s,\ccvec,\pi)} d\ccvec d\tilde{\theta} ds, \nonumber \\
    &\!=\!-\int_{s,\ccvec} q(s, \edt{C}|\pi) \Big \{\!\int_{\tilde{\theta}} q(\tilde{\theta}|s,\!\ccvec,\!\pi) \log \frac{q(\tilde{\theta}|s,\!\ccvec,\!\pi)}{q(\tilde{\theta}|\edt{s}, \pi)} d\tilde{\theta} \Big\} d\ccvec ds, \nonumber \\
    &\!=\!-\mathbb{E}_{q(s,\ccvec|\pi)} \Big[ D_{KL} \Big(q(\tilde{\theta}|s,\ccvec,\pi) || q(\tilde{\theta}|\edt{s}, \pi) \Big) \Big].
\end{align}
}By combining these parts, (\ref{eq: efe}) is finally derived. 
\vspace{-0.1in}

\section{Approximation for the First Term of (\ref{eq: efe_entropy})} \label{appendix: efe_pareto_derivation_first}
In order to approximate the first term of (\ref{eq: efe_entropy}), $\log p_{\pref}(\ccvec)$ can be at first simplified as follows. 
{\allowdisplaybreaks
\abovedisplayskip = 3.5pt
\abovedisplayshortskip = 3.5pt
\belowdisplayskip = 3.5pt
\belowdisplayshortskip = 1.5pt
\begin{align}
    \log p_{\pref}(\ccvec)\!&=\!\log\!\Big( Z_{\pref}\!\exp \!\big(\!-\!\frac{1}{2}(\ccvec\!-\!\mathbf{\mu}_{\pref})^T \mathbf{\Sigma}_{\pref}^{-1} (\ccvec\!-\!\mathbf{\mu}_{\pref})\big)\Big), \nonumber\\
    &= \log ( Z_{\pref})\!-\!\frac{1}{2} (\ccvec\!-\!\mathbf{\mu}_{\pref})^T \mathbf{\Sigma}_{\pref}^{-1} (\ccvec\!-\!\mathbf{\mu}_{\pref}) \big),
\end{align}
}where $Z_{\pref} = (2\pi)^{-N/2} |\mathbf{\Sigma}_{\pref}|^{-1/2}$.
Recall, the certainty equivalence approximation for the predicted observation distribution $q(\ccvec | \pi) \approx q(\ccvec | \pi ; \expec{\theta})$ is a MVN distribution $q(\ccvec | \pi) = \mathcal{N}(\mathbf{\mu}_{s, \pi}, \mathbf{\Sigma}_{s, \pi})$. By substituting this result into (\ref{eq: efe_entropy}), the first term of (\ref{eq: efe_entropy}) can be further transformed as 
\begin{align} \label{eq: efe_entropy_intrim}
    &(\mbox{1\textsuperscript{st}})\!=\!-\mathbb{E}_{q(\ccvec|\pi;\mathbb{E}[\theta])}\Big[\log(Z_{\pref})\!-\!\frac{(\ccvec\!-\!\mathbf{\mu}_{\pref})^T \mathbf{\Sigma}_{\pref}^{-1} (\ccvec\!-\!\mathbf{\mu}_{\pref})}{2}\Big].
\end{align}
Since $(\ccvec\!-\!\mathbf{\mu}_{\pref})^T \mathbf{\Sigma}_{\pref}^{-1} (\ccvec\!-\!\mathbf{\mu}_{\pref})$ is expanded as $\ccvec^T\mathbf{\Sigma}_{\pref}^{-1}\ccvec\!-\!2\mathbf{\mu}_{\pref}^T \mathbf{\Sigma}_{\pref}^{-1} \ccvec\!+\! \mathbf{\mu}_{\pref}^T\mathbf{\Sigma}_{\pref}^{-1}\mathbf{\mu}_{\pref}$, the second term of (\ref{eq: efe_entropy_intrim}) is turned to 
\begin{align}
    (\mbox{2\textsuperscript{nd}} \ \mbox{of} \ (\ref{eq: efe_entropy_intrim})) &= \frac{1}{2} \int_\ccvec q(\ccvec |\pi ; \expec{\theta}) \big( \ccvec^T \mathbf{\Sigma}_{\pref}^{-1} \ccvec \big) d\ccvec \label{eq:app_fta_1} \\
    & + \frac{1}{2} \int_\ccvec q(\ccvec | \pi ; \expec{\theta}) \big( \mathbf{\mu}_{\pref}^T \mathbf{\Sigma}_{\pref}^{-1} \mathbf{\mu}_{\pref} \big) d\ccvec \label{eq:app_fta_2} \\
    & - \int_\ccvec q(\ccvec | \pi ; \expec{\theta}) \big( - \mathbf{\mu}_{\pref}^T \mathbf{\Sigma}_{\pref}^{-1} \ccvec \big) d\ccvec \label{eq:app_fta_3}
\end{align}
Finally, (\ref{eq:app_fta_1}) can be reduced to
\begin{equation}
    \frac{1}{2} \Big( tr(\mathbf{\Sigma}_{\pref}^{-1} \mathbf{\Sigma}) + \mathbf{\mu}^T \mathbf{\Sigma}_{\pref}^{-1} \mathbf{\mu} \Big) \nonumber
\end{equation}
where $\expec{\theta} = \{\mathbf{\mu}, \mathbf{\Sigma}\}$. Also, (\ref{eq:app_fta_2}) is simply
\begin{equation}
    \mathbf{\mu}_{\pref}^T \mathbf{\Sigma}_{\pref}^{-1} \mathbf{\mu}_{\pref}, \nonumber
\end{equation}and (\ref{eq:app_fta_3}) can be reduced to
\begin{equation}
    - \mathbf{\mu}_{\pref}^T \mathbf{\Sigma}_{\pref}^{-1} \mathbf{\mu}. \nonumber
\end{equation}

\section{\edt{Empirical Analysis of Error Introduced via NIW-MVN Substitution}} \label{appendix: mvn_sub}
\edt{
To evaluate the choice of approximate posterior described in Sec. \ref{ssec: third_term}, we numerically demonstrated the efficacy of substituting intractable convolved NIW distributions (denoted $p(\cdot)$) with convolutions of MVN (denoted $q(\cdot)$).}  
\edt{Additionally, since the same substitution technique is used for choosing the prior proposal distribution in \eqref{eq: moment_match}, this analysis also yields insight into the quality of the variational upper bound.}

\begin{figure*}[t]
    \begin{subfigure}[t]{0.48\textwidth}
        \centering
        \includegraphics[trim={45mm 100mm 60mm 85mm},clip,width=1.0\textwidth]{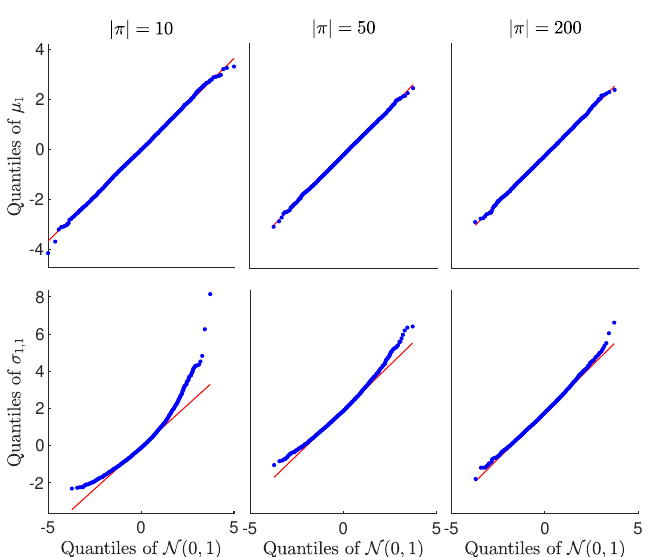}
        \caption{\edt{Q-Q plot varying length of plan ($m = |\pi|$), while each $(s,a)$ has 10 data samples ($l$). As the length of the plan increases, the true distribution becomes more normal (i.e. the effect of the CLT).}}
        \label{fig:plan_length_qq}
    \end{subfigure}
    ~~
    \begin{subfigure}[t]{0.48\textwidth}
        \centering
        \includegraphics[trim={45mm 100mm 60mm 85mm},clip,width=1.0\textwidth]{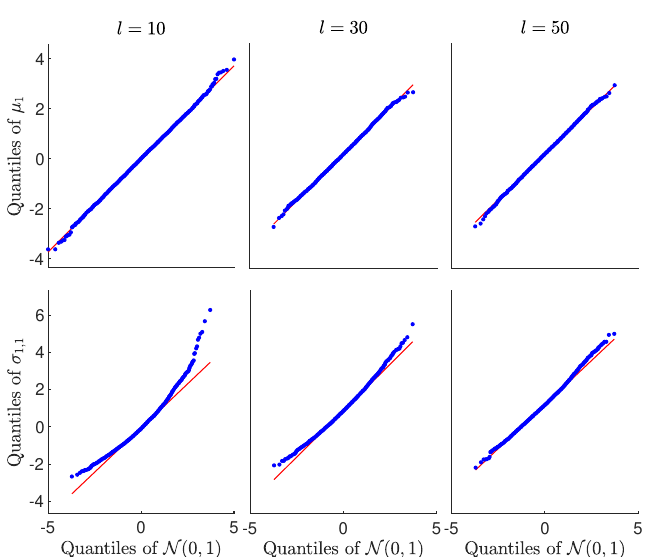}
        \caption{\edt{Q-Q plots of varying number of collected samples ($l$) from execution for each state-action, while $m=20$. As more samples are collected, the true distribution becomes more normal.}}
        \label{fig:data_qq}
    \end{subfigure}
    \caption{\edt{Q-Q Plot assessing the normality of Monte Carlo approximated true posterior $p(\tilde{\theta}|s, \ccvec, \pi)$. For brevity, only $\mu_1$ and $\sigma_{1,1}$ are shown. The closer the samples (blue) are to the quantile line (red), the more normal the empirical distribution.}}
    \label{fig:qq}
\end{figure*}
\begin{figure*}[t]
    \begin{subfigure}[t]{0.5\textwidth}
        \centering
        \includegraphics[trim={0mm 0mm 0mm 0mm},clip,width=1.0\textwidth]{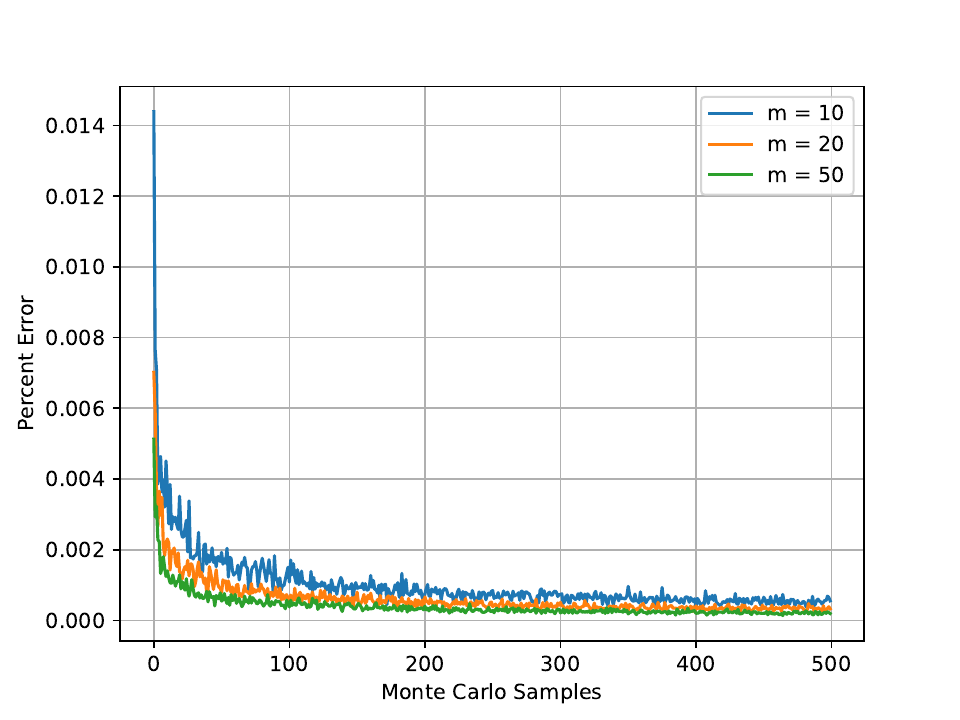}
        \caption{\edt{Error for varying length of plan ($m$).}}
        \label{fig:n_samples_m}
    \end{subfigure}
    \begin{subfigure}[t]{0.5\textwidth}
        \centering
        \includegraphics[trim={0mm 0mm 0mm 0mm},clip,width=1.0\textwidth]{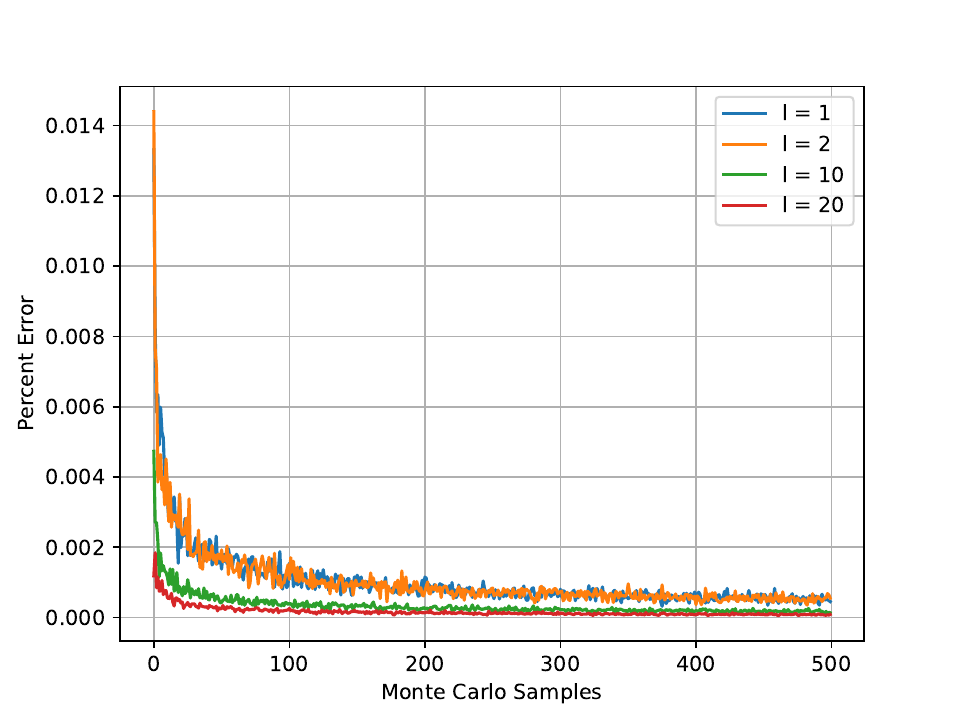}
        \caption{\edt{Error for varying number of collected samples ($l$).}}
        \label{fig:n_samples_l}
    \end{subfigure}
    \caption{\edt{Percent error introduced via sampling.}}
    \label{fig:n_samples}
\end{figure*}

\edt{Following \eqref{eq: expec_params_1} and \eqref{eq: expec_params_2}, $p(\cdot)$ and $q(\cdot)$ must have the same mean and variance, therefore, it suffices to analyze the normality of Monte Carlo samples of $p(\cdot)$ for determining how well $q(\cdot)$ approximates $p(\cdot)$. 
This analysis is done using a Quantile-Quantile (Q-Q) plot \cite{qq2017}.
The procedure for generating Monte Carlo samples used in the Q-Q plot is outlined as follows:
\begin{enumerate}
    \item Generate $M$ number of samples of $\hat{\tilde{\theta}}_{s,a}$ from each respective NIW $p(\tilde{\theta}_{s_k, a_k})$. For high fidelity, we used $M = 5000$ samples.
    \item Sum each respective sample to get $M$ samples of $\hat{\tilde{\theta}}_{s_K, \pi}$, (empirically representing samples from $p(\tilde{\theta}|s, \ccvec, \pi)$), 
    \item Perform a whitening transform to all $\hat{\tilde{\theta}}_{s_K, \pi}$ such that the such that the mean is zero and the covariance is normalized and uncorrelated,
    \item Produce a Q-Q plot to assess the normality of the whitened empirical distribution.
\end{enumerate}
The resulting Q-Q plots are shown in Fig. \ref{fig:qq}. The distributions of the mean parameters $\mu_{s_K, \pi}$ (Fig. \ref{fig:plan_length_qq}, top row) are generally well approximated by $q(\cdot)$. For short plans $|\pi| = 10$, $p(\cdot)$ the covariance parameter distributions differ from $q(\cdot)$ slightly (Fig. \ref{fig:plan_length_qq}, bottom left), however, due to the Central Limit Theorem (CLT), they converge to normal as the length of the plan increases (Fig. \ref{fig:plan_length_qq}, bottom right). Similarly, as more data is supplied to each state-action pair, the distribution becomes more normal due to the Inverse Wishart distribution being less right-skewed (Fig. \ref{fig:data_qq}, bottom left to bottom right). Therefore, the MVN replacement approximation introduces less error as the scenarios become more complex, and more data is collected. 
}

\section{\edt{Empirical Analysis of Error Introduced via Monte Carlo Sampling}} \label{appendix: mc_sampling}
\edt{Fig. \ref{fig:n_samples} shows the percent error introduced via sampling. Note that $n_s=300$ was used for the experiments and benchmarks shown in the manuscript as well as the computation time benchmarks.
Although the specific sampling procedure is slightly different, there are multiple studies using the Monte Carlo sampling to evaluate the EFE \cite{fountas2020deep, tschantz2020, Maisto2021ActiveTS}.}

\end{document}